\documentclass{article}
\pdfoutput=1
\usepackage{arxiv}

\usepackage[utf8]{inputenc} 
\usepackage[T1]{fontenc}    
\usepackage{url}            
\usepackage{booktabs}       
\usepackage{amsfonts}       
\usepackage{nicefrac}       
\usepackage{microtype}      
\usepackage{lipsum}
\usepackage{graphicx}
\usepackage{subfigure}
\usepackage{amsmath}
\usepackage{amssymb}
\usepackage{amsthm}
\usepackage{algorithm}
\usepackage{algorithmic}
\usepackage{hyperref}

\DeclareMathOperator*{\argmax}{arg\,max}

\DeclareMathOperator{\score}{score}

\usepackage{placeins}

\usepackage{hyperref}

\begin{document}
  \title{Clustered Hierarchical Anomaly and Outlier Detection Algorithms}

\author{
    Najib Ishaq \\
    Department of Computer Science and Statistics\\
    University of Rhode Island\\
    Kingston, RI\\
    \texttt{najib\_ishaq@uri.edu} \\
    \And
    Thomas J. Howard III \\
    Department of Computer Science and Statistics\\
    University of Rhode Island\\
    Kingston, RI\\
    \texttt{thoward27@uri.edu} \\
    \AND
    Noah M. Daniels \\
    Department of Computer Science and Statistics\\
    University of Rhode Island\\
    Kingston, RI\\
    \texttt{noah\_daniels@uri.edu} \\
}

    \maketitle

  \begin{abstract}
    Anomaly and outlier detection is a long-standing problem in machine learning.
    In some cases, anomaly detection is easy, such as when data are drawn from well-characterized distributions such as the Gaussian.
    However, when data occupy high-dimensional spaces, anomaly detection becomes more difficult.
    We present CLAM (Clustered Learning of Approximate Manifolds), a manifold mapping technique in any metric space.
    CLAM begins with a fast hierarchical clustering technique and then induces a graph from the cluster tree, based on overlapping clusters as selected using several geometric and topological features.
    Using these graphs, we implement CHAODA (Clustered Hierarchical Anomaly and Outlier Detection Algorithms), exploring various properties of the graphs and their constituent clusters to find outliers.
    CHAODA employs a form of transfer learning based on a training set of datasets, and applies this knowledge to a separate test set of datasets of different cardinalities, dimensionalities, and domains.
    On 24 publicly available datasets, we compare CHAODA (by measure of ROC AUC) to a variety of state-of-the-art unsupervised anomaly-detection algorithms.
    Six of the datasets are used for training.
    CHAODA outperforms other approaches on 16 of the remaining 18 datasets.
    CLAM and CHAODA scale to large, high-dimensional ``big data'' anomaly-detection problems, and generalize across datasets and distance functions.
    Source code to CLAM and CHAODA are freely available on GitHub\footnote{https://github.com/URI-ABD/clam}.
  \end{abstract}

    \section{Introduction and Related Work}
\label{sec:introduction}

Detecting anomalies and outliers from data is a well-studied problem in machine learning.
When data occupy easily-characterizable distributions, such as the Gaussian, the task is relatively easy:
one need only identify when a datum is sufficiently far from the mean.
However, in ``big data'' scenarios, where data can occupy high-dimensional spaces, anomalous behavior becomes harder to quantify.
If the data happen to be uniformly distributed, one can conceive of simple mechanisms, such as a one-class SVM, that would be effective in any number of dimensions.
However, real-world data are rarely distributed uniformly.
Instead, data often obey the ``manifold hypothesis''~\cite{fefferman2016testing}, occupying a low-dimensional manifold in a high-dimensional embedding space, similar to how a 2-d sheet of paper, once crumpled, occupies a 3-dimensional space.
Detecting anomalies in such a landscape is not easy.
Imagine trying to identify if a point within the sphere of crumpled paper were anomalous; identifying whether or not the point sits on the sheet or within a gap of the lower-dimensional manifold presents a challenge.

Anomalies (data that do not belong to a distribution) and outliers (data which represent the extrema of a distribution) are found in datasets for a variety of reasons.
In this manuscript, we treat anomalies and outliers interchangeably.
Errors in the measurement of data create anomalous records,
novel instances of data emerge over time,
or normal behavior may evolve towards the outliers.
Moreover, even if data collection is executed perfectly and shifts in behavior are accounted for, there is still the risk of adversarial attacks causing undesired behavior~\cite{elsayed2018adversarial}.

Modern algorithms designed to detect anomalous behavior fail for a variety of reasons, in particular when anomalies live close to, but not on, a complex manifold in high-dimensional space.
Our approach is designed to map these complex manifolds.
Here we briefly survey contemporary approaches to anomaly detection in order to provide the context needed to understand how our approach differs.

\subsection{Clustering-based Approaches}
\label{subsec:introduction:clustering-based-approaches}

Clustering refers to techniques for grouping points in a way that provides value.
This is generally done by assigning \textit{similar} points to the same cluster.
Given a clustering and a new point, one can estimate the anomalousness of that new point by measuring its distance to its nearest cluster(s).

There have been few advancements in clustering techniques over the past decade~\cite{wang2019progress}.
This may be explained by the poor performance, thus far, of clustering in high-dimensional space~\cite{zhang2013advancements}.

Distance-based clustering relies on some distance measure to partition data into several clusters.
Within this approach, the numbers or sizes of clusters are often predetermined: either user-specified, or randomly chosen~\cite{wang2019progress}.
Examples of distance-based clustering algorithms are
K-Means~\cite{macqueen1967some},
PAM~\cite{kaufman2009finding},
CLARANS~\cite{ng1994efficient} and
CLARA~\cite{kaufman2009finding}.

Hierarchical clustering methods build a tree, where points are allocated into leaves~\cite{wang2019progress}.
These trees can be created via bottom-up (agglomerative) or top-down (divisive)~\cite{agrawal1998automatic} clustering.
The high cost of all-pairs distance computations is often a drawback.
Examples of hierarchical clustering algorithms include
MST~\cite{zahn1971graph},
CURE~\cite{guha1998cure}, and
CHAMELEON~\cite{karypis1999hierarchical}.

Grid-based clustering works via segmenting the entire volume into a discrete number of cells, and then scanning those cells to find regions of high density.
The grid structure typically lets these methods scale well to larger datasets.
Some examples include
STING~\cite{wang1997sting},
Wavecluster~\cite{sheikholeslami2000wavecluster},
CLIQUE~\cite{agrawal1998automatic},
Clustering Based Local Outlier Factor (CBLOF)~\cite{he2003cblof}, and
Local Correlation Integral (LOCI)~\cite{papadimitriou2003loci}.

\subsection{Density-based Approaches}
\label{subsec:introduction:density-based-approaches}

Density-based clustering methods rely on finding regions of high point-density separated by regions of lower density.
These algorithms generally do not work well when data are sparse or uniformly distributed.
Some examples include
DBSCAN~\cite{ester1996density} and
DENCLUE~\cite{hinneburg1998efficient}.
In this paper, we compare against Local Outlier Factor (LOF)~\cite{breunig2000lof}.

\subsection{Graph-based Approaches}
\label{subsec:introduction:graph-based-approaches}

Graph-based methods build graph representations of the data by treating points, or collections of points, as nodes in a graph and drawing edges between them based on various rules.
One challenge with graph-based methods is finding a suitable graph representation for a dataset; a user-specified distance threshold is often used to induce edges.
In this paper, we compare against Connectivity-based Outlier Factor (COF)~\cite{tang2002cof}.

\subsection{Distance-based Approaches}
\label{subsec:related-works:distance-based-approaches}

Distance-based methods use direct distance comparisons and largely rely on k-Nearest Neighbors as their substrate~\cite{wang2019progress}.
They tend to use the following intuitions.
First, points with fewer than $p$ other points within some distance $d$ are outliers;
second, the $n$ points with the greatest distances to their $k^{th}-$nearest neighbor are outliers;
and finally, the $n$ points with the greatest average distance to their $k$ nearest neighbors are outliers.
Drawbacks to distance-based approaches include the computational complexity of the underlying KNN algorithm, and determining a good value for $k$.
We compare against
k-Nearest Neighbors (kNN)~\cite{ramaswamy2000efficient, fabrizio2002knn} and
Subspace Outlier Detection (SOD)~\cite{kriegel2009sod}.

\subsection{Deep Learning Approaches}
\label{subsec:related-works:deep-learning-approaches}
With the recent explosion in Deep Learning, several new methods for anomaly detection have been introduced~\cite{pang2021deep}.
While most such approaches are supervised or weakly-supervised, such as DAGMM~\cite{zong2018deep}, there have been some advances in unsupervised methods.
These typically use autoencoders~\cite{chen2017outlier}, such as RandNet~\cite{chandola2009anomaly}, or generative-adversarial-networks, such as MO-GAAL and SO-GAAL~\cite{liu2019generative}.
Drawbacks to deep learning approaches include interpretability of the model, the wide variety of possible architecture and thus hyperparameters, and intensive computational and GPU requirements during training.
We compare against MO-GAAL, SO-GAAL, and two autoencoders from~\cite{chen2017outlier}.

\subsection{Other Approaches}
\label{subsec:introduction:other-appraoches}

There are several approaches to anomaly detection that do not fall neatly into any of the aforementioned categories.
These methods often rely on support vector machines, random forests, or histograms to detect outliers.
We compare against seven methods among these:
Histogram-Based Outlier Detection (HBOS)\cite{goldstein2012histogram},
Isolation-Forest (IFOREST)~\cite{tony2008iforest,tony2012iforest},
One-class Support Vector Machine (OCSVM)~\cite{sholkopf2001ocsvm},
Linear Model Deviation-base outlier Detection (LMDD)~\cite{arning1996linear},
Lightweight Online Detector of Anomalies (LODA)~\cite{pevny2016loda},
Minimum Covariance Determinant (MCD)~\cite{rousseeuw1999mcd,hardin2004mcd}, and
Subspace Outlier Detection (SOD)~\cite{kriegel2009sod}.

\subsection{Our Approach}
\label{subsec:introduction:chaoda}

With the term \emph{manifold learning} being largely synonymous with dimension reduction, we propose \emph{manifold mapping} to refer to approaches that study the the geometric and topological properties of manifolds in their original embedding spaces.
We introduce a novel technique, Clustered Learning of Approximate Manifolds (CLAM) for datasets in a metric space (more general than a Banach space);~\cite{banach1929fonctionnelles} essentially a set of datapoints and a distance metric defined on that set.
CLAM presupposes the manifold hypothesis and uses a divisive hierarchical clustering to build a map of the manifold occupied by the dataset.
CLAM then provides this map of the manifold to be used by a collection of anomaly detection algorithms, which we call CHAODA (Clustered Hierarchical Anomaly and Outlier Detection Algorithms).

CLAM extends CHESS~\cite{ishaq2019clustered} by adding memoized calculations of several geometric and topological properties of clusters that are useful to CHAODA, and does so in expected $\mathcal{O}(n \lg n)$ time.
While, in principle, we could have used any hierarchical clustering algorithm, these memoized calculations are not provided for by any other algorithm.
Other clustering algorithms also suffer from problems, such as:
an ineffective treatment of high dimensionality,
an inability to interpret results, and
an inability to scale to exponentially-growing datasets~\cite{agrawal1998automatic}.
CLAM, as we will demonstrate, largely resolves these issues.
Thus, CLAM and CHAODA are well-suited to anomaly detection on large, high-dimensional ``big data.''

CHESS was used to build a hierarchical clustering to a user-specific depth for the sole purpose of accelerating search.
CLAM, however, divisively clusters the data until each cluster contains only one datum.
Using the cluster-tree, CLAM induces graphs by mapping specific clusters to vertices of a graph, and drawing an edge between any two vertices whose corresponding clusters have overlapping volumes (i.e., the distance between their centers is less than or equal to the sum of their radii).
Clusters can be selected from a fixed depth, or from heterogeneous depths based on properties such as their local fractal dimension, cardinality, radius, etc.
We can consider clusters at lower depths in the tree to be, in some sense, ``lower resolution'' than those at greater depths.
Inducing a graph across a variety of depths effectively maps a manifold with a variety of ``resolutions,'' the intuition being that some regions of the manifold may have a higher density of points than others and, thus, graphs induced from clusters deeper in the tree may be more informative for those regions.

Having mapped a manifold by clustering and inducing graphs, we can start to analyze several properties of the clusters in the graphs.
For example: what are the relative cardinalities of the clusters in the graph, how connected are the clusters, how often is each cluster visited by a random walk?
CHAODA uses answers to such questions, among others, to build an ensemble approach to anomaly detection.

    \section{Methods}
\label{sec:methods}

\begin{figure*}[ht!]
    \centering
    \includegraphics[width=6in]{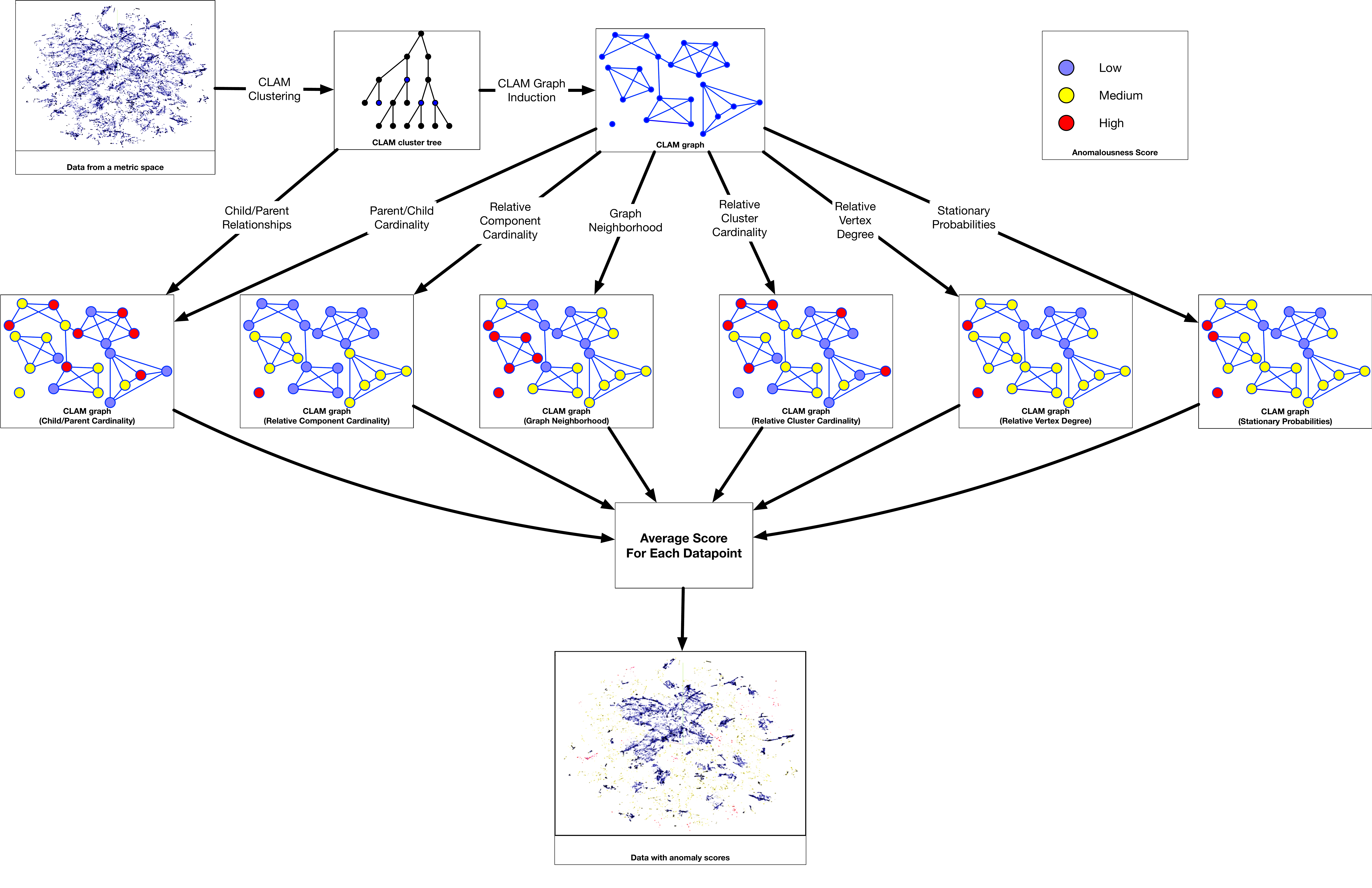}
    \caption{\textbf{Overview of the CHAODA workflow.}
        Beginning with a dataset and a distance metric, CLAM builds a cluster tree and induces several graphs from this tree; for the sake of simplicity, we illustrate only one such graph here.
        Each of CHAODA's constituent algorithms provides distinct anomaly scores on its graph.
        These scores are normalized and aggregated into a final score for each cluster, and by extension, each datum.
        In this figure, we have simplified the scores to a ternary color scheme; actual scores are real-valued between 0 and 1.
        Note that each algorithm provides its own scoring, but there may be similarities such as between vertex degree and stationary distribution.}
    \label{fig:methods:chaoda-workflow}
\end{figure*}

CLAM and CHAODA comprise many components, all described in this section.
We start with a brief overview of those components.
CLAM begins with a dataset and a distance metric, to which it applies hierarchical clustering to build a tree.
CLAM selects clusters from the tree using meta-machine-learning (meta-ml) models trained\footnote{Note that during training, CHAODA is somewhat supervised.
CHAODA uses this training to learn a set of meta-ml models for selecting clusters and inducing graphs.
During testing, and inference, on new datasets, CHAODA is unsupervised and uses the learned meta-ml models.} according to several geometric and topological properties.
These meta-ml models learn relationships between these properties and expected anomaly-detection performance.
CLAM then induces graphs from the selected clusters.
CHAODA applies its constituent algorithms to these graphs, and combines the individual scores into an ensemble, ultimately producing anomaly scores for each datum.
See Figure~\ref{fig:methods:chaoda-workflow} for a high-level illustration.

\subsection{Dataset and Distance Function}
\label{subsec:methods:dataset-and-distance-function}

We start with a dataset $\textbf{X} = \{x_1 \dots x_n\}$ with $n$ points and a distance function $f : (\textbf{X}, \textbf{X}) \mapsto \mathbb{R}^+$.
The distance function takes two points from the dataset and deterministically produces a non-negative real number.
We also require the distance function to be symmetric and for the distance between two identical points to be zero, i.e., $\forall x, y \in X$, $f(x, y) = f(y, x)$ and $f(x, y) = 0 \Leftrightarrow x = y$.
CLAM and CHAODA are general over any distance function that obeys these constraints.

CLAM assumes the ``manifold hypothesis''~\cite{fefferman2016testing}, i.e.\ datasets collected from constrained generating phenomena that are embedded in a high-dimensional space typically only occupy a low-dimensional manifold in that space.
CLAM and CHAODA learn the geometric and topological properties of these manifolds in a way that generalizes across datasets and distance function regardless of dataset-specific properties such as total number of points, dimensionality, absolute distance values, etc.
We demonstrate this genericity by our anomaly detection performance in Section~\ref{sec:results}.

Note that we often speak of each datum as embedded in a $D$-dimensional metric space and we use Euclidean notions, such as voids and volumes, to talk about the geometric and topological properties of the manifold.
The purpose of such notions is to help build intuition and to aid understanding.
Mathematically, CLAM does not rely on such notions; in-fact, the details of an embedding space can be abstracted away behind the distance function.

Also note that we can provide certain guarantees (see CHESS~\cite{ishaq2019clustered}) when the distance function is a metric, i.e.\ it obeys the triangle inequality.
While CLAM and CHAODA work well with distance functions that are not metrics, we have not investigated how the lack of the triangle inequality changes, or breaks, those guarantees in the context of anomaly detection.
For this paper, we show results using the $L1$-norm and $L2$-norm.

\subsection{Clustering}
\label{subsec:methods:clustering}

We start by building a divisive hierarchical clustering of the dataset.
We partition, as described in Algorithm~\ref{alg:partition}, a cluster with $k$ points using a pair of well-separated points from among a random sampling of $\sqrt k$ points.
Starting from a root-cluster containing the entire dataset, we continue until each leaf contains only one datum.
This achieves clustering in expected $\mathcal{O}(n \lg n)$ time.
This procedure improves upon the clustering approach from CHESS~\cite{ishaq2019clustered} by a better selection of maximally-separated points, and by memoizing critical information about each cluster (discussed below).

\begin{algorithm} 
\caption{Partition} 
\label{alg:partition} 
\begin{algorithmic}[1] 
    \REQUIRE $cluster$
    \STATE $k \leftarrow \lfloor \sqrt{|cluster.points|} \rfloor$
    \STATE $seeds \leftarrow k$ random points from $cluster.points$
    \STATE $c \leftarrow$ geometric median of $seeds$
    \STATE $r \leftarrow \argmax d(c,x) \ \forall \ x \in cluster.points$
    \STATE $l \leftarrow \argmax d(r,x) \ \forall \ x \in cluster.points$
    \STATE $left \leftarrow \{x | x \in cluster.points \land d(l,x) \le d(r,x)\}$
    \STATE $right \leftarrow \{x | x \in cluster.points \land d(r,x) < d(l,x)\}$
    \IF{$|left| > 1$}
        \STATE Partition($left$)
    \ENDIF
    \IF{$|right| > 1$}
        \STATE Partition($right$)
    \ENDIF
\end{algorithmic}
\end{algorithm}

These clusters have several interesting and important properties for us to consider.
These include the \textit{cardinality}, the number of points in a cluster;
\textit{center}, the approximate geometric median of points contained in a cluster;
\textit{radius}, the distance to the farthest point from the center;
and \textit{local fractal dimension},
as given by:

\begin{gather}
    \log_2\bigg(\frac{|B_X(c, r)|}{|B_X(c, \frac{r}{2})|}\bigg)
    \label{fractal-dimension}
\end{gather}

where $B_X(c,r)$ is the set of points contained in a ball of radius $r$ on the dataset $X$ centered on a point $c$~\cite{ishaq2019clustered}.
Thus, local fractal dimension captures the ``spread'' of points on the manifold in comparison to the (typically much larger) embedding space.
This is motivated by the idea that the induced graphs will learn to adapt to use different ``resolutions'' to characterize different regions of the manifold (see Figure~\ref{fig:methods:cluster-resolution}).

We can also consider \textit{child-parent ratios} of the cardinality, radius, and local fractal dimension of a cluster, as well as the \textit{exponential moving averages} of those child-parent ratios along a branch of the tree.
In particular, we use the child-parent ratios and their exponential moving averages to help CHAODA generalize from a small set of training datasets to a large, distinct set of testing datasets.
During clustering, we memoize these ratios as we create each cluster.
CHAODA can then make direct use of these ratios to aid in anomaly detection.

\subsection{Graphs}
\label{subsec:methods:graphs}

Clusters that are close together sometimes have overlapping volumes; i.e.,\ the distance between their centers is less than or equal to the sum of their radii.
We define a graph $G=(V,E)$ with vertices in one-to-one correspondence to CLAM clusters and with an edge between two vertices if and only if their corresponding clusters overlap.
While it is fairly standard in the literature to define graphs in this way, the challenge lies in selecting the right clusters to build useful graphs.
Our selection process, presented in Section~\ref{subsec:methods:cluster-selection-for-graphs}, is among the major novel contributions of CLAM and CHAODA.

In the context of graphs, we use the terms \textit{cluster} and \textit{vertex} interchangeably.
By \textit{graph cardinality} we mean \textit{vertex cardinality}, i.e., the number of clusters in the graph, and by \textit{graph population} we mean the sum of cardinalities of all clusters in the graph.
Note that \textit{cluster cardinality} refers to the number of points within a cluster.
We use \textit{layer-graph} to refer to a graph built from clusters at a fixed depth from the tree and \textit{optimal-graph} to refer to a graph built from clusters selected by the processes described in Section~\ref{subsec:methods:cluster-selection-for-graphs}.

\begin{figure}[ht!]
    \centering
    \includegraphics[width=2.5in]{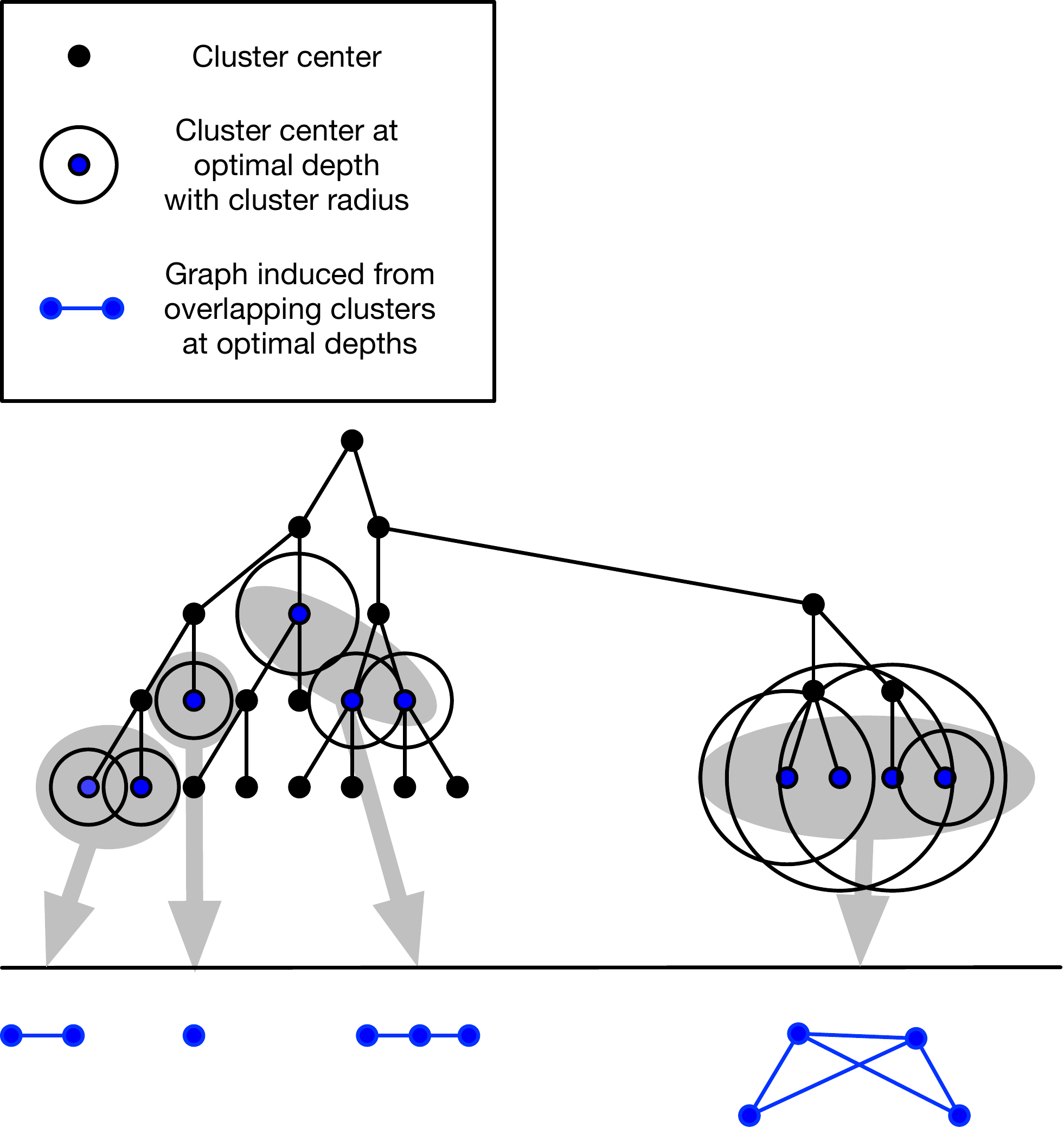}
    \caption{Using CLAM to induce a graph from a cluster tree.
        Dots in the tree represent cluster centers;
        blue dots represent centers of chosen clusters.
        Circles represent the volume of a cluster (the radius is the distance from the center to the furthest point contained within that cluster).
        Gray arrows point to the induced graph components, which are indicated in blue below the horizontal line.}
    \label{fig:methods:graph-generation}
\end{figure}

\begin{figure}[ht!]
    \centering
    \includegraphics[width=2.5in]{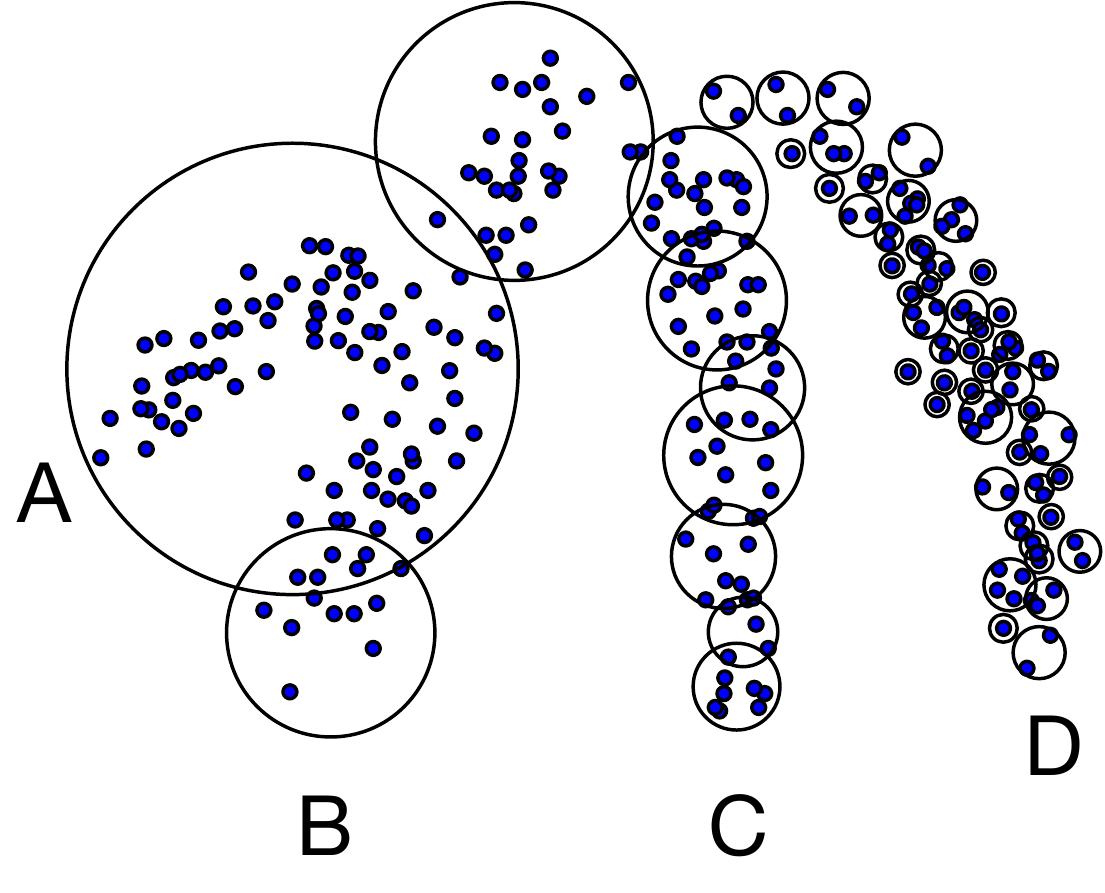}\\
    \vspace{1cm}
    \includegraphics[width=2.5in]{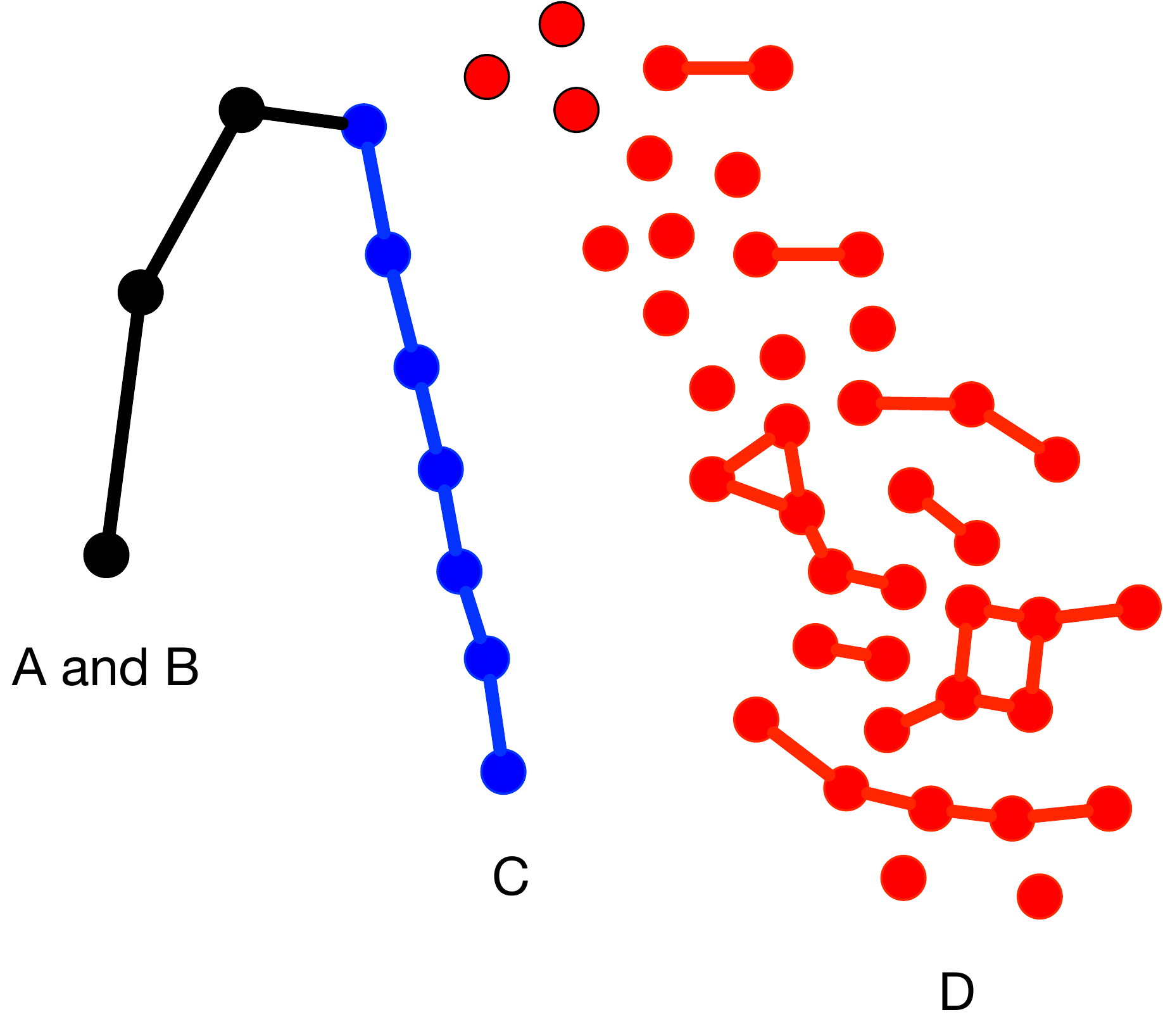}
    \caption{
        \textbf{Cluster resolution.}
        Consider this toy dataset whose manifold comprises the four named branches.
        In the top part of the figure, with each branch, we show clusters that we might get from different depths in the tree.
        Clusters on branches A and B come from low depths in the tree and have large voids, i.e.\ regions with no points in them.
        Clusters on branch D come from a high depth in the tree and are, in a sense, too small for the ``thickness'' of the branch they cover.
        Clusters on branch C are ``just right'' because their diameters are roughly equal to the thickness of the branch, and they contain no large voids.
        We can track how the local fractal dimension of these clusters changes as we traverse the tree and as we move along clusters that are adjacent on the manifold.
        In this way, changes in the local fractal dimension can serve as a proxy for deciding which clusters would help ``well characterize'' the underlying manifold.
        In the bottom part of the figure, we show the graphs CLAM would induce from these different clusters.
        Note that branches A and B are not distinguished; the separation between the branches is lost in the graph representation.
        A graph induced from branch D would consist of many disconnected subgraphs, and would not represent the structure of the entire branch.
        Finally, a graph induced from branch C represents the branch structure, including its connection to branches A and B.
    }
    \label{fig:methods:cluster-resolution}
\end{figure}

Figure~\ref{fig:methods:graph-generation} illustrates how CLAM induces a graph from non-uniform depths in a cluster tree, while Figure~\ref{fig:methods:cluster-resolution} illustrates how, if clusters are chosen at the right ``resolution,'' these graphs can capture the structure of the manifold.
Interestingly, the clusters are not necessarily hyperspheres, but polytopes akin to a high-dimensional Voronoi diagram~\cite{voronoi1908nouvelles}.
The induced graph need not be fully connected and, in practice, often contains many small, disjoint connected components.

For our purposes, a CLAM graph exhibits an important invariant.
The clusters corresponding to vertices in the graph collectively contain every point in the dataset, and each point in the dataset is assigned to exactly one cluster in the graph.
A corollary to this invariant is that a graph will never contain two clusters such that one cluster is an ancestor or descendant of another cluster.
This also assures that a \textit{graph's population} is equal to the cardinality of the dataset, i.e. $|\textbf{X}|$ or $n$.

\subsection{Individual Algorithms}
\label{subsec:methods:individual-algorithms}

Given an induced graph that characterizes a manifold, we must extract information about the anomalousness of clusters in that graph.
Here we describe six simple algorithms for anomaly detection, each using a CLAM graph to calculate an anomalousness score for each cluster and datum.
Given that the key to an effective ensemble is for each member to contribute a unique inductive bias~\cite{chen2017outlier}, we also note the intuition behind each algorithm's contributions.
These scores can be used, along with the ground-truth labels, to compute the area under the curve (AUC) of the receiver operating characteristic (ROC)~\cite{fawcett2006introduction} to measure the anomaly detection performance of the graph which produced those scores.

In the following,
$V$ and $E$ are the sets of clusters and edges respectively in a graph,
$|c|$ is the cardinality of a cluster $c$,
and $|C|$ is the cardinality of a component $C$.
Each algorithm assigns an anomalousness score to each cluster.
Each point is then assigned the anomalousness score of the cluster it belongs to.
These scores are internally consistent for each individual algorithm, i.e.\ low scores indicate inliers and high scores indicate outliers.
However, different algorithms assign scores in wide, and often different, ranges of values.
We use Gaussian normalization to constrain the raw scores to a $[0, 1]$ range.
This lets us combine scores into an ensemble (see Section~\ref{subsec:methods:the-ensemble}).
See~\cite{kriegel2011interpreting} for a thorough discussion of anomaly-score normalization in ensemble methods.

The overall computational complexity of these algorithms appears in Table~\ref{table:methods:complexity}.

\begin{table}[ht!]
\centering
\caption{Time complexity of CHAODA algorithms.}
\label{table:methods:complexity}
\begin{tabular}{|l|l|}
\hline
\textbf{Algorithm} & \textbf{Complexity} \\
\hline
 Relative Cluster Cardinality & $\mathcal{O}(|V|)$ \\
 \hline
 Relative Component Cardinality & $\mathcal{O}(|E| + |V|)$  \\
 \hline
 Graph Neighborhood Size & $\mathcal{O}(|E| \cdot |V|)$  \\ 
 \hline
 Child-Parent Cardinality Ratio & $\mathcal{O}(|V|)$ \\ 
 \hline
 Stationary Probabilities & $\mathcal{O}(|V|^{2.37})$ \\ 
 \hline
 Relative Vertex Degree & $\mathcal{O}(|V|)$ \\
 \hline
\end{tabular}
\end{table}

\subsubsection{Relative Cluster Cardinality}
\label{subsubsec:methods:individual-algorithms:relative-cluster-cardinality}
We measure the anomalousness of a point by the cardinality of the cluster that the point belongs to relative to the cardinalities of the other clusters in the graph.
Points in the same cluster are considered equally anomalous.
Points in clusters with lower cardinalities are considered more anomalous than points in clusters with higher cardinalities.
Formally, $\forall c \in G, \ \score(c) = -|c|$.

The intuition is that points in clusters with higher cardinalities are close to each other, and thus are less likely to be anomalous.
The time complexity is $\mathcal{O}(|V|)$ because this requires a single pass over the clusters in a graph.

\subsubsection{Relative Component Cardinality}
\label{subsubsec:methods:individual-algorithms:relative-component-cardinality}
We use the usual definition of connected components:
no two vertices from different components have an edge between them and every pair of vertices in the same component has a path connecting them.
We consider points in clusters in smaller components to be more anomalous than points in clusters in larger components.
Points in clusters in the same component are considered equally anomalous.
Formally, $\forall C \in G, \ \forall c \in C,  \\score(c) = -|C|$.

The intuition here, as distinct from the previous algorithm, is to capture larger-scale structural information based on disjoint connected components from the graph.
The time complexity is $\mathcal{O}(|E| + |V|)$ because we first need to find the components of the graph using a single pass over the edges, and then score each cluster in the graph using a single pass over those clusters.

\subsubsection{Graph Neighborhood Size}
\label{subsubsec:methods:individual-algorithms:graph-neighborhood-size}
Given the graph, we consider the number of clusters reachable from a starting cluster within a given graph distance $k$, i.e.\ within $k$ hops along edges.
We call this number the \textit{graph-neighborhood size} of the starting cluster.
With $k$ small compared to the diameter of a component, we consider the relative graph-neighborhood-sizes of all clusters.
Clusters with small graph-neighborhoods are considered more anomalous than clusters with large graph-neighborhoods.

The intuition here is to capture information about the connectivity of the graph in the region around each cluster.
The computation is defined in Algorithm~\ref{alg:graph-neighborhood-size}.
Its time complexity is $\mathcal{O}(|E| \cdot |V|)$ because we need to compute the eccentricity of each cluster.

\begin{algorithm}[h]
    \caption{Graph Neighborhood}
    \label{alg:graph-neighborhood-size}
\begin{algorithmic}[1]
    \REQUIRE $G$, a graph
    \REQUIRE $\alpha \in \mathbb{R}$ in the range $(0,1]$ ($0.25$ by default).
    \FOR {cluster $c \in G$}
        \STATE $e_c \gets$ the eccentricity of $c$
        \STATE $s \gets e_c \cdot \alpha$  
        \STATE perform a breadth-first traversal from $c$ with $s$ steps
        \STATE $v \gets$ the number of unique clusters visited
        \STATE $\score(c) \gets -v$
    \ENDFOR
\end{algorithmic}
\end{algorithm}

\subsubsection{Child-Parent Cardinality Ratio}
\label{subsubsec:methods:individual-algorithms:child-parent-cardinality-ratio}
As described in Section~\ref{subsec:methods:clustering}, the partition algorithm used in clustering splits a cluster into two children.
If a child cluster contains only a small fraction of its parent's points, then we consider that child cluster to be more anomalous.
These child-parent cardinality ratios are accumulated along each branch in the tree, terminating when the child cluster is among those selected in the graph.
Clusters with a low value of these accumulated ratios are considered more anomalous than clusters with a higher value.
Formally, $\forall c \in G, \ \score(c) = \frac{|p|}{|c|} + score(p)$ where $p$ is the parent cluster of $c$.

This algorithm was inspired by iForest~\cite{tony2008iforest}, and captures information from the tree and the graph.
Unlike the other individual algorithms, this accumulates parent scores into the children.
The time complexity of this algorithm is $\mathcal{O}(|V|)$, because these ratios are memoized during the clustering process and we need only look them up once for each cluster in the graph.

\subsubsection{Stationary Probabilities}
\label{subsubsec:methods:individual-algorithms:stationary-probabilities}
For each edge in the graph, we assign a weight inversely proportional to the distance between the centers of the two clusters that connect to form that edge.
The outgoing probabilities from a cluster are stochastic over the edge weights for that cluster.
We compute the transition probability matrix of each component that contains at least two clusters.
The process of successively squaring this matrix will converge~\cite{levin2017markov}.
We follow this process for each component in the graph and find the convergent matrix.
Consider the sum of the values along a row in this matrix.
This is the expected proportion of visits to that cluster during an infinitely long random walk over the component.
We consider this sum to be inversely related to the anomalousness of the corresponding cluster.

The intuition here is that clusters that are more difficult to reach during an infinite random walk are more likely to contain anomalous points.
The algorithm is defined in Algorithm~\ref{alg:stationary-probabilities}.
Its worst-case time complexity is $\mathcal{O}(|V|^{2.37})$ given by the matrix multiplication algorithm from~\cite{alman2021refined}.
In practice, however, this algorithm has much better performance than indicated by the theoretical complexity, because the induced graphs are often composed of several small components rather than one, or a few large, component(s).

\begin{algorithm}[h]
    \caption{Stationary Probabilities}
    \label{alg:stationary-probabilities}
\begin{algorithmic}[1]
    \REQUIRE $G$, a graph
    \FOR {component $C \in G$}
        \STATE $M \gets$ the transition matrix for $C$
        \REPEAT
            \STATE $M \gets M^2$
        \UNTIL $M$ converges
        \FOR {cluster $c \in C$}
            \STATE $s \gets $ the row from $M$ corresponding to $c$
            \STATE $\score(c) \gets -\Sigma(s)$ 
        \ENDFOR
    \ENDFOR
\end{algorithmic}
\end{algorithm}

\subsubsection{Relative Vertex Degree}
\label{subsubsec:methods:individual-algorithms:relative-vertex-degree}
For each cluster in the induced graph, consider its degree, i.e. the number of edges connecting to that cluster.
We consider a cluster with high degree to be less anomalous than a cluster with low degree.
This is essentially a version of the previous algorithm that ignores edge weights, and will have different biases with regard to the sampling density of the dataset.
Formally, $\forall c \in G, \ \score(c) = -\deg(c)$.
Its time complexity is $\mathcal{O}(|V|)$.

\subsection{Training Meta-Machine-Learning Models}
\label{subsec:methods:training-meta-ml-models}

Section~\ref{subsec:methods:clustering} makes note of some important geometric and topological properties of CLAM clusters, i.e.\ cardinality, radius and local fractal dimension.
We find the child-parent ratios of these properties and the exponential moving averages of these ratios along each branch in the tree.
Each child-parent ratio is obtained by dividing the value for the child by the value for the parent, e.g.\ $R= \frac{|child|}{|parent|}$.
Each new exponential moving average (EMA) is the weighted sum of the previous EMA and the current ratio.
Specifically, $ema_{i+1} = \alpha * R_{i + 1} + (1 - \alpha) * ema_i$ for some $\alpha \in [0, 1]$.
We chose an $\alpha$ of $\frac{2}{11}$.

Using these ratios instead of the raw values themselves makes CHAODA agnostic to dataset-specific properties; it need only consider how those properties change as we traverse the tree or a graph.
For a given graph, we can take the average values of the six ratios from its constituent clusters to form a feature-vector.
We can use the methods described in Section~\ref{subsec:methods:individual-algorithms} to compute the area under the ROC curve from using each individual algorithm to predict anomalousness scores from that graph.
Each pairing of the feature-vector and an ROC score forms a training sample for our meta-ml models.
We use linear regression and decision-tree regressors to fill the role of those meta-ml models.
We use these data to train the meta-ml models to predict the ROC score for a graph from its feature-vector.

We randomly selected six datasets whose cardinalities are between $10^3$ and $10^5$ for training, and we used the $L1$-norm and $L2$-norm for each dataset.
For each pairing of dataset and distance function, CLAM builds a new cluster-tree.
Meta-ml training then proceeds over several epochs, the first of which we seed with some layer graphs from each tree.
During each epoch, we extract the feature vector from each graph, and we find the ROC AUC of applying each individual algorithm to each graph.
Each pairing of feature-vector and ROC score forms a training sample.
For each pairing of dataset and distance function, we initialize a linear regressor and a decision-tree regressor to form our suite of meta-ml models.
We train each meta-ml model with every training sample collected thus far, for ten epochs.
We use the trained meta-ml models to select clusters (see Section~\ref{subsec:methods:cluster-selection-for-graphs}) for new graphs that are used for the next epoch.
We note that this was not $k$-fold cross validation, but a one-time selection of six datasets for training based on size as a selection criterion.

During the earlier epochs, we expect to have selected graphs that exhibit poor anomaly detection performance.
For later epochs, we expect this performance to improve.
With each epoch, we add to the set of training samples collected thus far and we train a new suite of meta-ml models for selecting better clusters.
This is so that the meta-ml models can learn to distinguish between ratios that select for low ROC AUC from those ratios that select for high ROC AUC.
Each meta-ml model sees training data from each pairing of dataset and distance function.
This lets CHAODA generalize across different datasets and distance functions.

\subsection{Cluster Selection for Graphs}
\label{subsec:methods:cluster-selection-for-graphs}

The heart of the problem with CHAODA is in selecting the ``right'' clusters that would build a graph that provides a useful representation of the underlying manifold.
One could try every possible combination of clusters to build graphs, but this quickly leads to combinatorial explosion.
Instead, CHAODA focuses on intelligently selecting clusters for a graph which is expected to perform well for anomaly detection.
Area under the curve (AUC) of the receiver operating characteristic (ROC) is often used to benchmark anomaly detectors~\cite{fawcett2006introduction}.
CHAODA selects clusters to optimize for this measure.

\begin{figure}[ht!]
    \centering
    \includegraphics[width=2.5in]{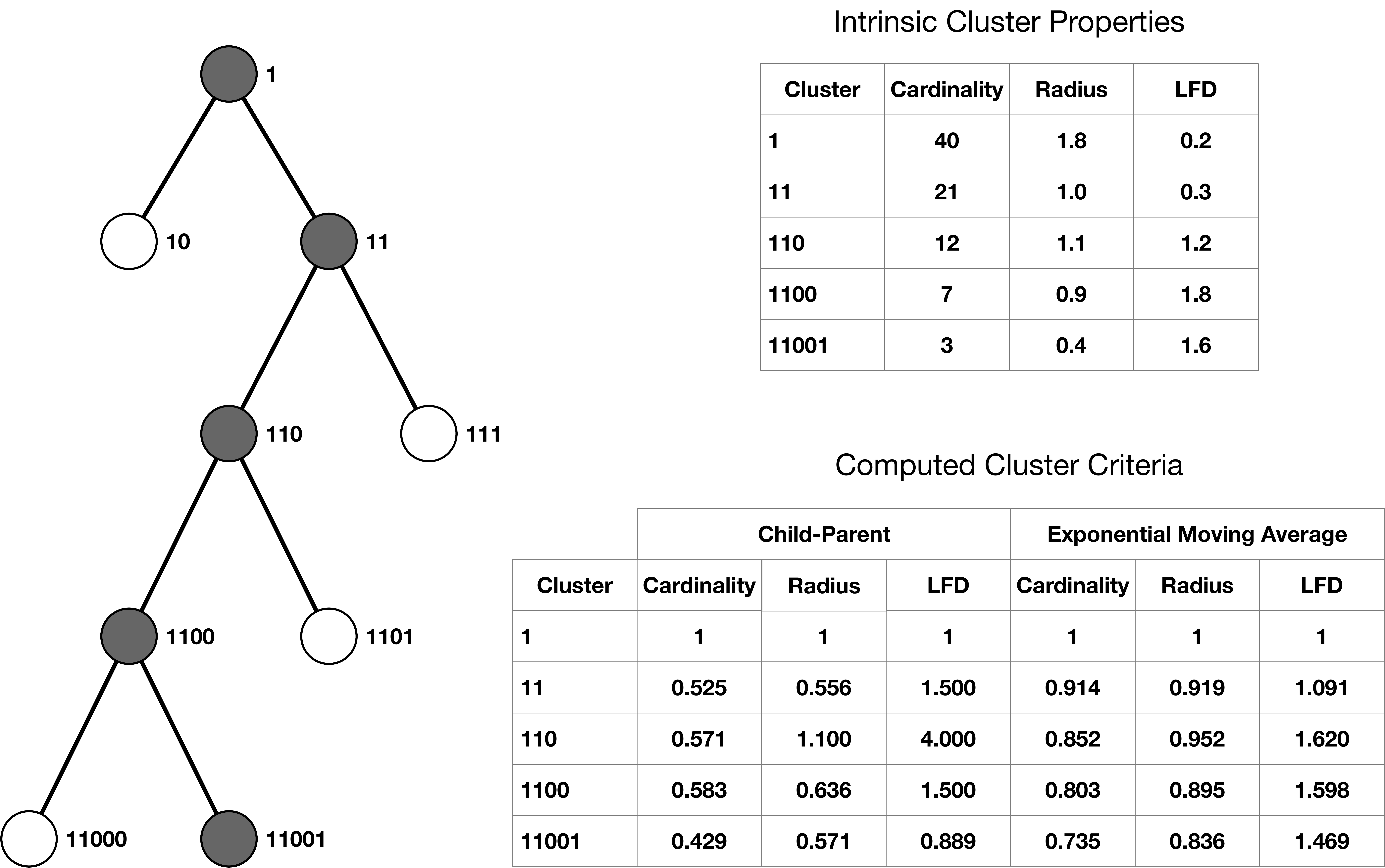}
    \caption{
        \textbf{Cluster properties.}
        In the illustrated tree, we highlight only one branch for simplicity.
        We name the root `1' and we name the descendants as we might for a huffman tree.
        The upper table is an example of the values that intrinsic cluster properties might take on.
        The lower table shows the derived ratios we use for learning how to select clusters.
    }
    \label{fig:methods:cluster-properties}
\end{figure}

Specifically, we train a number of meta-ml models (see Section~\ref{subsec:methods:training-meta-ml-models} for details) and, from each model, we extract a function of the form $g : c \rightarrow \mathbb{R}$.
This function assigns high values to clusters which would increase ROC AUC and low values to clusters which would decrease ROC AUC.
As described in Algorithm~\ref{alg:cluster-selection}, the selection process begins by sorting, in non-increasing order, all clusters in the tree by the value assigned by $g$.
This sorting represents a ranking of the clusters for expected anomaly detection performance.
We iteratively select the best cluster from the rankings, and with each selection, we remove the ancestors and descendants of the selected cluster from the list of rankings.
Once the list of rankings is exhausted, we have selected the clusters with which to build an \textit{optimal graph}.

\begin{algorithm}[h]
    \caption{Cluster Selection}
    \label{alg:cluster-selection}
\begin{algorithmic}[1]
    \REQUIRE $T$, a cluster-tree
    \REQUIRE $g : c \rightarrow \mathbb{R}$ a ranking function
    \STATE $G \gets$ an empty graph
    \STATE $h \gets$ a list of all clusters $c \in T$ sorted by $g(c)$
    \REPEAT
        \STATE $c \gets$ pop\_first($h$)
        \STATE Add $c$ to $G$
        \STATE Remove all ancestors and descendants of $c$ from $h$
    \UNTIL $h$ is empty
\end{algorithmic}
\end{algorithm}

\subsection{The Ensemble}
\label{subsec:methods:the-ensemble}

During the testing and inference phases, we begin with a new dataset (not included in the training set of datasets) and one or more distance functions.
CLAM first builds cluster-trees using each distance function with the given dataset.
CHAODA uses the trained meta-ml models to select a different graph from each tree for each individual algorithm.
CHAODA applies each individual algorithm to its corresponding graph and produces anomalousness scores for each datum.
With two distance functions, six individual algorithms, and two meta-ml models, we can get up to $24$ different members with which to form an ensemble.
CHAODA normalizes the scores from all members and aggregates them, by their mean, into a final set of predictions for the anomalousness of each datum.

\subsection{Datasets and Comparisons}
\label{subsec:methods:datasets-and-comparisons}

We sourced 24 datasets containing only numerical features, i.e.\ not categorical features, from Outlier Detection Datasets (ODDS)~\cite{rayana2016odds}.
All of these datasets were adapted from the UCI Machine Learning Repository (UCIMLR)~\cite{UCIMLR}, and were standardized by ODDS for anomaly detection benchmarks.
Note that CHAODA is able to handle either entirely-numerical or entirely-categorical datasets, but not mixed datasets.
We discuss some future work relating to this in Section~\ref{sec:discussion}.

We randomly selected six datasets to train CHAODA: ann-thyroid, mnist, pendigits, satellite, shuttle, and thyroid.
The other eighteen datasets were used for testing and benchmarks: arrhythmia, breastw, cardio, cover, glass, http, ionosphere, lymphography, mammography, musk, optdigits, pima, satimage-2, smtp, vertebral, vowels, wbc, and wine.
We benchmarked CHAODA 30 times, using different random seeds, on the test set of datasets (see the Supplement at https://github.com/URI-ABD/chaoda for more details).
During testing, we noticed that even though we often see $|V| \ll n $, the graph neighborhood size and stationary probabilities methods from~\ref{subsec:methods:individual-algorithms} took prohibitively long to run, so we only use them when $|V| < max(128, \lfloor \sqrt n \rfloor)$.
We present these results in Table~\ref{table:results:test-performance} under the CHAODA-fast and CHAODA rows.
CHAODA-fast exhibits comparable performance to CHAODA, and we offer it as an option in our implementation.
All benchmarks were conducted on a 28-core Intel Xeon E5-2690 v4 2.60GHz, 512GB RAM and CentOS 7 Linux with kernel 3.10.0-1127.13.1.el7.x86\_64 \#1 SMP and Python 3.6.8.

We use the ground-truth labels only during the training phase with a small set of datasets.
Having once been trained, CHAODA becomes an unsupervised algorithm for any new dataset.
As such, we compared CHAODA only against other unsupervised algorithms.
We selected $18$ unsupervised algorithms from the pyOD suite~\cite{zhao2019pyod} and Scikit-Learn~\cite{pedregosa2011scikit}, as well as RS-Hash~\cite{sathe2016subspace}.
A supervised version of CHAODA is possible future work, which would open up comparisons against supervised or weakly-supervised methods such as REPEN~\cite{pang2018learning} and DAGMM~\cite{zong2018deep}.

For a ``Big-Data'' challenge, we ran CHAODA on the APOGEE2 data from the SDSS~\cite{blanton2017sdss}.
This dataset has a cardinality of $528,319$ and a dimensionality of $8,575$.
See Section~\ref{subsec:results:sdss-apogee2} for results.
All of these datasets were prepared by UCI and ODDS. 
In our experiments, we simply read them as 2-dimensional arrays where the columns are the features and the rows are the instances. 
We pre-processed the APOGEE2 data into a similar array, but of course it has no ground-truth labeling.

    \section{Results}
\label{sec:results}

The performance on the 18 test datasets is in Table~\ref{table:results:test-performance}.
Performance on the 6 training datasets is shown in the Supplement at \url{https://homepage.cs.uri.edu/~ndaniels/pdfs/chaoda-supplement.pdf}.
Each column shows the ROC scores of CHAODA and every competitor.
The highest score and every score within $0.02$ is presented in bold.
We found that setting a different random seed resulted in a variance of at most $0.02$ ROC AUC for CHAODA.

If a method exceeded 10 hours on a dataset, we mark the corresponding cell with ``\textit{TO}''.
If a method crashed, we mark the cell with ``\textit{EX}''.
Notably, CHAODA performed best (or tied for best) on 16 of the 18 test datasets.
Runtime performance is presented in the Supplement. 
Note that we implemented CLAM and CHAODA entirely in Python, while the methods we compared against are often implemented in C/C++.
Therefore, the comparison of runtime is not truly fair to CHAODA.
An implementation in a high-performance language, such as Rust, would be worthwhile.

\begin{table*}[!t]
\caption{Performance (ROC AUC) of CHAODA vs. other methods on the 18 test datasets.}
\label{table:results:test-performance}
\vskip 0.15in
\begin{center}
\begin{small}
\begin{sc}
\begin{tabular}{|c|c|c|c|c|c|c|c|c|c|}
\hline
\textbf{Model} & \textbf{Arrh} & \textbf{BreastW} & \textbf{Cardio} & \textbf{Cover} & \textbf{Glass} & \textbf{Http} & \textbf{Iono.} & \textbf{Lympho} & \textbf{Mammo} \\
\hline
CHAODA-fast    & \textbf{0.79} &             0.78 &   \textbf{0.86} &           0.58 & \textbf{0.80}  & \textbf{0.99} &           0.72 &   \textbf{0.96} &  \textbf{0.85} \\
\hline
CHAODA         &          0.76 &    \textbf{0.94} &            0.82 &  \textbf{0.82} &           0.71 & \textbf{1.00} &  \textbf{0.88} &   \textbf{0.99} &  \textbf{0.86} \\
\hline
ABOD           &          0.62 &             0.50 &            0.49 &           0.51 &           0.53 &          0.50 &           0.85 &            0.80 &           0.50 \\
\hline
AutoEncoder    &          0.65 &             0.91 &            0.74 &           0.52 &           0.54 &          0.51 &           0.65 &            0.83 &           0.51 \\
\hline
CBLOF          &          0.70 &             0.83 &            0.57 &    \textit{EX} &           0.54 &   \textit{EX} &           0.86 &            0.83 &           0.50 \\
\hline
COF            &          0.65 &             0.26 &            0.50 &           0.50 &           0.59 &          0.51 &           0.81 &            0.83 &           0.51 \\
\hline
HBOS           &          0.65 &             0.93 &            0.58 &           0.49 &           0.48 &          0.51 &           0.36 &            0.91 &           0.50 \\
\hline
IFOREST        &          0.72 &             0.91 &            0.69 &           0.50 &           0.54 &          0.53 &           0.77 &            0.83 &           0.59 \\
\hline
KNN            &          0.68 &             0.84 &            0.51 &           0.51 &           0.54 &          0.51 &  \textbf{0.90} &            0.83 &           0.51 \\
\hline
LMDD           &          0.68 &             0.64 &            0.60 &           0.49 &           0.54 &          0.51 &           0.67 &            0.65 &           0.56 \\
\hline
LOCI           &          0.62 &      \textit{TO} &     \textit{TO} &    \textit{TO} &           0.58 &   \textit{TO} &           0.58 &            0.90 &    \textit{TO} \\
\hline
LODA           &          0.65 &             0.93 &            0.60 &           0.52 &           0.48 &          0.51 &           0.63 &            0.48 &           0.52 \\
\hline
LOF            &          0.67 &             0.30 &            0.49 &           0.50 &           0.54 &          0.51 &           0.79 &            0.83 &           0.53 \\
\hline
MCD            &          0.65 &             0.94 &            0.55 &           0.50 &           0.48 &          0.50 &  \textbf{0.90} &            0.83 &           0.51 \\
\hline
MOGAAL         &          0.42 &             0.40 &            0.45 &    \textit{TO} &           0.59 &   \textit{TO} &           0.36 &            0.48 &    \textit{TO} \\
\hline
OCSVM          &          0.70 &             0.77 &            0.70 &           0.56 &           0.54 &          0.50 &           0.68 &            0.83 &           0.60 \\
\hline
SOD            &          0.59 &             0.77 &            0.48 &    \textit{TO} &           0.54 &   \textit{TO} &           0.84 &            0.65 &           0.51 \\
\hline
SOGAAL         &          0.48 &             0.30 &            0.45 &           0.61 &           0.59 &          0.51 &           0.36 &            0.48 &           0.50 \\
\hline
SOS            &          0.51 &             0.50 &            0.50 &    \textit{TO} &           0.48 &   \textit{TO} &           0.72 &            0.48 &    \textit{TO} \\
\hline
VAE            &          0.65 &    \textbf{0.95} &            0.74 &           0.52 &           0.48 &          0.51 &           0.65 &            0.83 &           0.56 \\
\hline
\hline
\textbf{Model} & \textbf{Musk} & \textbf{OptDigits} & \textbf{Pima} & \textbf{SatImg-2} & \textbf{Smtp} & \textbf{Vert} & \textbf{Vowels} & \textbf{WBC}  & \textbf{Wine} \\
\hline
CHAODA-fast    & \textbf{1.00} &               0.57 &          0.57 &     \textbf{0.98} & \textbf{0.96} &          0.29 &   \textbf{0.83} &          0.93 & \textbf{1.00} \\
\hline
CHAODA         & \textbf{1.00} &      \textbf{0.96} &          0.60 &     \textbf{1.00} & \textbf{0.95} &          0.29 &   \textbf{0.90} & \textbf{0.97} & \textbf{0.99} \\
\hline
ABOD           &          0.47 &               0.54 &          0.60 &              0.53 &          0.50 &          0.49 &            0.75 &          0.50 &          0.43 \\
\hline
AutoEncoder    &          0.63 &               0.48 &          0.57 &              0.71 &          0.50 &          0.49 &            0.51 &          0.77 &          0.51 \\
\hline
CBLOF          & \textbf{1.00} &               0.52 & \textbf{0.64} &              0.90 &          0.50 &          0.49 &            0.52 &          0.82 &          0.46 \\
\hline
COF            &          0.53 &               0.52 &          0.54 &              0.56 &          0.50 &          0.51 &            0.71 &          0.47 &          0.46 \\
\hline
HBOS           & \textbf{1.00} &               0.60 &          0.55 &              0.49 &          0.68 &          0.47 &            0.56 &          0.77 &          0.57 \\
\hline
IFOREST        &          0.97 &               0.50 & \textbf{0.65} &              0.94 &          0.50 &          0.45 &            0.63 &          0.72 &          0.51 \\
\hline
KNN            &          0.51 &               0.51 &          0.60 &              0.61 &          0.53 &          0.47 &            0.72 &          0.51 &          0.47 \\
\hline
LMDD           &          0.48 &               0.49 &          0.37 &              0.49 &          0.65 &          0.43 &            0.49 &          0.80 &          0.62 \\
\hline
LOCI           &   \textit{TO} &        \textit{TO} &   \textit{TO} &       \textit{TO} &   \textit{TO} &          0.49 &     \textit{TO} &          0.72 &          0.46 \\
\hline
LODA           &          0.54 &               0.51 &          0.62 &              0.69 &          0.57 &          0.43 &            0.51 &          0.82 &          0.57 \\
\hline
LOF            &          0.50 &               0.53 &          0.55 &              0.55 &          0.50 &          0.49 &            0.69 &          0.50 &          0.46 \\
\hline
MCD            &          0.97 &               0.48 & \textbf{0.66} &              0.61 &          0.50 &          0.45 &            0.63 &          0.60 &          0.46 \\
\hline
MOGAAL         &          0.48 &               0.48 &          0.61 &              0.49 &   \textit{TO} &          0.51 &            0.48 &          0.60 &          0.46 \\
\hline
OCSVM          &          0.48 &               0.49 &          0.56 &              0.84 &          0.50 &          0.49 &            0.50 &          0.82 &          0.46 \\
\hline
SOD            &          0.51 &               0.51 &          0.56 &              0.58 &   \textit{TO} &          0.45 &            0.66 &          0.60 &          0.46 \\
\hline
SOGAAL         &          0.48 &               0.52 &          0.48 &              0.49 &          0.62 & \textbf{0.54} &            0.48 &          0.47 &          0.46 \\
\hline
SOS            &          0.52 &               0.52 &          0.51 &              0.52 &   \textit{TO} &          0.49 &            0.59 &          0.52 &          0.46 \\
\hline
VAE            &          0.63 &               0.48 &          0.61 &              0.71 &          0.50 &          0.45 &            0.51 &          0.77 &          0.67 \\
\hline
\end{tabular}
\end{sc}
\end{small}
\end{center}
\vskip -0.1in
\end{table*}

We considered several recently published algorithms against which to compare.
Those with available implementations are included in Table~\ref{table:results:test-performance}.
When unable to find a working implementation, we include here the performance claimed by the respective authors.
RS-Hash~\cite{sathe2016subspace} reported AUCs of $0.92$ on Cardio, $1.00$ on Lympho, $0.99$ on Musk, and $0.76$ on OptDigits.
This beats CHAODA on Cardio, ties on Lympho and Musk, and is outperformed by CHAODA on OptDigits.
We considered Clustering with Outlier Removal~\cite{liu2019clustering} but we could not find a working implementation, and the authors did not report AUC scores, instead only reporting F-measure.
We considered comparisons against REPEN~\cite{pang2018learning} and RDP~\cite{wang2019unsupervised}, but REPEN's publicly available source code lacks information about dependencies and their versions, and training RDP took prohibitively long.

\subsection{SDSS-APOGEE2}
\label{subsec:results:sdss-apogee2}

We demonstrate the ability of CLAM and CHAODA to scale to the next generation of ``Big-Data'' problems.
As a proof of concept, we consider the APOGEE2 data.
This dataset contains spectra of a large number of stars collected, so far, during the SDSS project~\cite{blanton2017sdss}.
We extracted $528,319$ spectra in $8,575$ dimensions and used CHAODA, under the $L1$-norm and the $L2$-norm, to produce anomaly scores.
Since there is no ground-truth available, we simply report the scores and the time taken in the Supplement.

\section{UMAP Visualization}
\label{sec:umap-visualization}

\begin{figure*}
   \centering
   \includegraphics[width=2.1in]{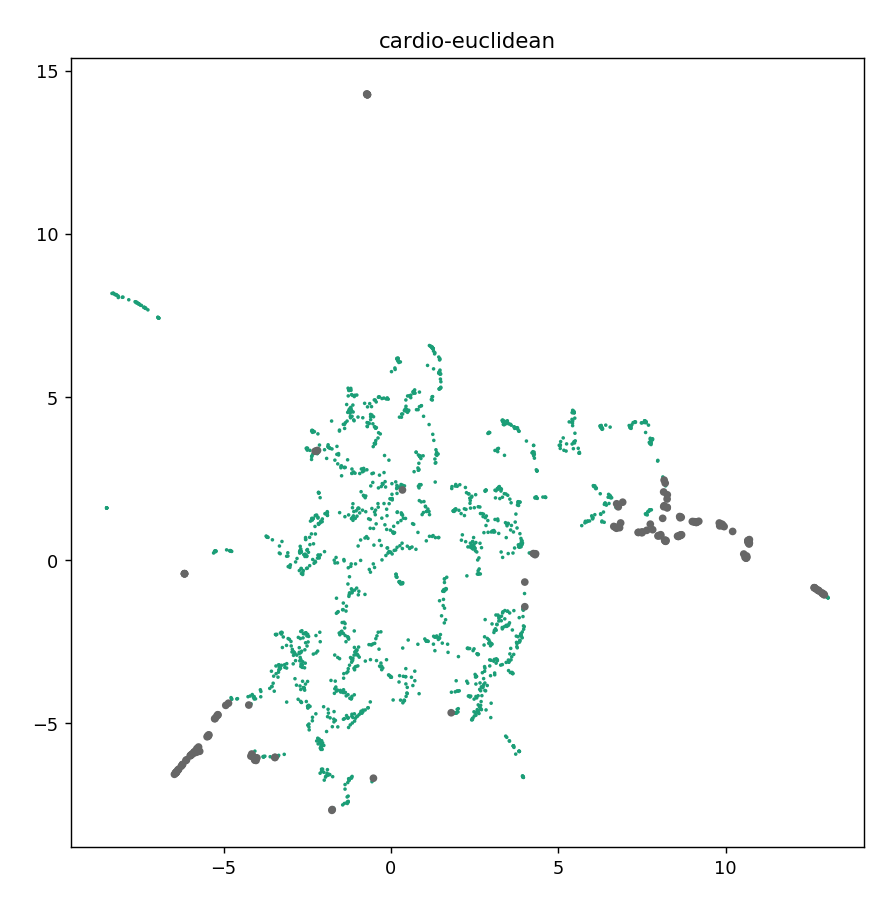}
   \includegraphics[width=2.1in]{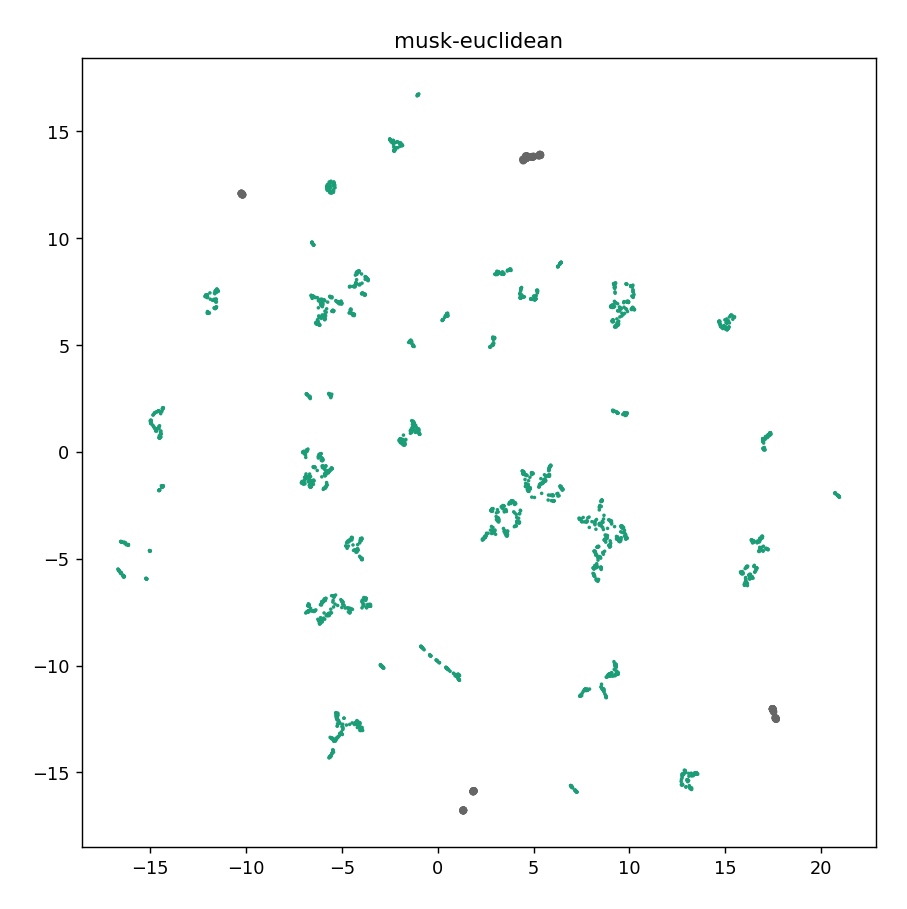}
   \includegraphics[width=2.1in]{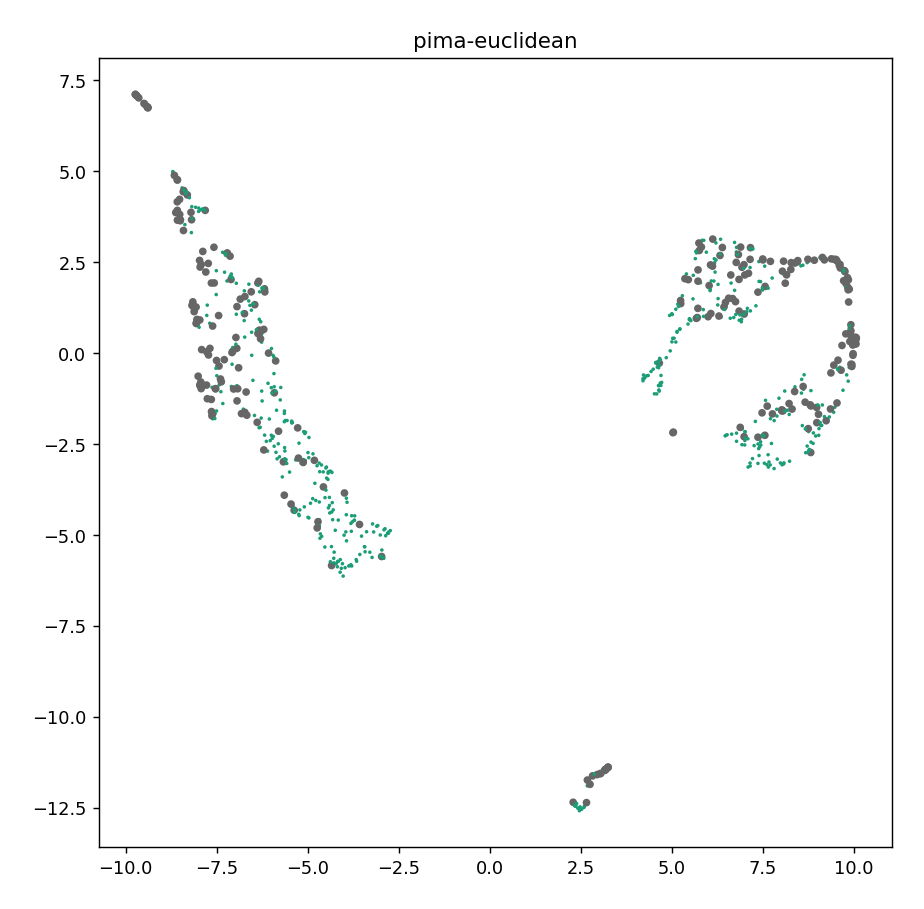}
   \caption{UMAP projections of Cardio (left), Musk (middle), and Pima (right) under Euclidean distance.
   For Cardio, there is a single main component to the manifold, and anomalies tend to be at the edges of that manifold.
   For Musk, there are several distinct pieces to the manifold, perhaps corresponding to different digits.
   CHAODA outperforms other approaches on Cardio, while many approaches achieve perfect performance on Musk. On Pima, all approaches fare poorly, and the UMAP projection illustrates that the anomalies cover much of the manifold, including in the interior.}
   \label{fig:conclusions:umap-embeddings-1}
\end{figure*}

A visualization in Figure~\ref{fig:conclusions:umap-embeddings-1} using UMAP illustrates three different examples;
the anomalies in the Cardio dataset, where CHAODA outperforms other methods, appear to be at the edges of a complex manifold (though, clearly, the UMAP projection has distorted the manifold). In the Musk dataset, where many methods including CHAODA achieve perfect performance, there are several distinct components to the manifold, likely corresponding to different digits.
In the Pima dataset, all methods perform fairly poorly, the anomalies appear to be distributed across the manifold, including in the interior.

    \section{Discussion}
\label{sec:discussion}

We have presented CHAODA, an ensemble of six algorithms that use the map of the underlying manifold produced by CLAM\@.
The six individual algorithms are simple to implement on top of CLAM and, when combined into an ensemble, often outperform state-of-the-art methods.
In future work, new ideas for additional algorithms that contribute novel inductive biases can be easily incorporated into CHAODA.

While the meta-ml models in CHAODA optimize for the best ROC-AUC scores for anomaly detection, this approach can be used to optimize for any measure on any type of task.
Future work should explore tasks other than anomaly detection and come up with mathematical measures for performance on those tasks.
Variants of CHAODA could then be trained for such tasks.

Cross-validation on a single dataset is commonplace in machine learning.
While $k$-fold cross-validation splits a single dataset into $k$ folds and then evaluates performance on each fold separately, these folds are still drawn from the same underlying distribution; indeed, this is the principle that justifies cross-validation.
In contrast, CHAODA's meta-ml approach learns geometric and topological properties for inducing graphs from a \emph{set} of training datasets, and \emph{transfers} that knowledge to an \emph{entirely distinct} set of datasets, which may differ in cardinality, dimensionality, domain, and the generating phenomenon or process that produced the data.

CLAM uses the geometric and topological properties such as fractal dimension of the data to build a map of the low-dimensional manifold that the data occupy.
CLAM extends CHESS~\cite{ishaq2019clustered} by:
a better selection of ``poles'' (i.e. a pair of well-separated points) for partitioning clusters,
memoizing important cluster properties, and
introducing a novel graph-induction approach using a notion of optimal depths, learned via a form of ``meta-machine-learning'' and transfer learning.
Whereas CHESS was developed specifically for accelerating search,
CHAODA uses this manifold-mapping framework to discover properties of the manifold that are useful for anomaly detection.
Intuitively, we expect CHAODA to perform particularly well when the data lie on an ``interesting'' manifold, and to perform merely average when the data derive from an easily-described distribution (or ``boring'' manifold).
Just as CHESS demonstrated an acceleration of search when the data exhibited \emph{low fractal dimension} and \emph{low metric entropy}, we see that CHAODA vastly improves ROC scores when the data exhibit these properties.
When the data do not exhibit these properties, CHAODA is still often competitive with other approaches.
CLAM is free of hyper-parameters other than the fairly standard choice of $\alpha$ in Section~\ref{subsec:methods:training-meta-ml-models}; the weights learned from the meta-ml step could vary, but we learned them once on a distinct training set of datasets.

We briefly discuss the Vertebral Column (Vert.) dataset, with regards to the robustness of our approach.
On this dataset, most algorithms we evaluated performed similarly to random guessing, while CHAODA performed much worse.
We suspect this is due to how this specific dataset was collected.
Each instance represents six biomechanical attributes derived from scans of a patient's pelvis and lumbar spine.
This dataset contains 210 instances of the Abnormal class treated as inliers and 30 instances of the Normal class treated as outliers.
Each attribute must have a narrow range to be in the Normal class, but can have a wider range in the Abnormal class.
This causes the Normal instances to group together, while Abnormal instances remain distant from each other.
As CHAODA relies on clusters as the substrate, it assigns low scores to instances in the Normal class, i.e.\ the outliers, and high scores to those in the Abnormal class, i.e.\ the inliers.
Put plainly, CHAODA sees the manifold as the Normal class, which the ground-truth labels as outliers.

The choice of distance function could significantly impact anomaly-detection performance.
In this case, domain knowledge is likely the best way to determine the distance function of choice.
Future work should explore a more diverse collection of domain-appropriate distance functions, such as Wasserstein distance on images, Levenshtein distance on strings, and Jaccard distance on the maximal common subgraph of molecular structures.
Currently, CLAM only functions on a metric space defined by a distance metric (it is not, however, limited to complete Banach spaces).
This poses a limitation on datasets that have heterogenous features, such as a mix of continuous and categorical variables.
Future work should explore linear combinations of normalized distance functions to overcome this limitation.
Additionally, we do not yet know how CHAODA would generalize across distance functions; i.e. predicting anomalousness with distance functions different from those used to train the meta-ml models.
Future work should investigate this question of generalization.
It would also be worth exploring the question of whether CHAODA extends to nondeterministic distance functions, as well as performance on other non-metric distance functions, such as those disobeying the triangle inequality (e.g. cosine distance or Damerau-Levenshtein edit distance~\cite{damerau1964technique}).

In this paper, we have used CHAODA (and the methods under comparison) to score entire datasets with known anomaly labels for purposes of evaluating CHAODA's accuracy.
In real-world usage, one might wish to assign anomaly scores to an incoming data stream.
This is a simple extension: given some corpus of data (some of which may or may not be anomalous), build a CLAM tree and the induced graphs, and assign anomaly scores from the CHAODA algorithms as we have demonstrated.
Then, as each new datum arrives, simply fit it into the CLAM tree ($\mathcal{O}(\lg |V|)$ time using tree-search from CHESS) into a cluster that is found in a graph and assign it the anomaly score for that cluster.
If an incoming datum is too far from any cluster (further than any existing datum at that depth from its cluster center) then it can initialize a new cluster, which would be assigned a high anomaly score.
Thus, in general, CHAODA requires $\mathcal{O}(\lg |V|)$ time to assign an anomaly score to a new datum.


CHAODA is demonstrably highly effective on large high-dimensional datasets, and so may be applied to neural networks.
Using CLAM to map a dataset where each datum represents the activation-pattern of a neural network from an input to the neural network, we would expect to detect malicious inputs to neural networks based on the intuition that malicious inputs produce atypical activation patterns.

In conclusion, we have demonstrated that by mapping the manifolds occupied by data, CLAM reveals geometric and topological structure that allows CHAODA to outperform other state-of-the-art approaches to anomaly detection, representing an actualization of the manifold hypothesis.

Supplementary results and figures are available at \url{https://homepage.cs.uri.edu/~ndaniels/pdfs/chaoda-supplement.pdf}.
The source code for CLAM and CHAODA are available under an MIT license at \url{https://github.com/URI-ABD/clam}.


    \FloatBarrier
    \bibliographystyle{acm}
    \bibliography{references}
    \newpage








\renewcommand{\thetable}{S\arabic{table}}

\renewcommand{\thefigure}{S\arabic{figure}}

\section{Datasets}
\label{supplement:sec:datasets}

Here we describe the datasets we use for benchmarks.
See Table~\ref{supplement:table:datasets} for a summary of this information.

The \textbf{annthyroid} dataset is derived from the ``Thyroid Disease'' dataset from the UCIMLR\@.
The original data has 7200 instances with 15 categorical attributes and 6 real-valued attributes.
The class labels are ``normal'', ``hypothyroid'', and ``subnormal''.
For anomaly detection, the ``hypothyroid'' and ``subnormal'' classes are combined into 534 outlier instances, and only the 6 real-valued attributes are used.

The \textbf{arrhythmia} dataset is derived from the ``Arrhythmia'' dataset from the UCIMLR\@.
The original dataset contains 452 instances with 279 attributes.
There are five categorical attributes which are discarded, leaving this as a 274-dimensional dataset.
The instances are divided into 16 classes.
The eight smallest classes collectively contain 66 instances and are combined into the outlier class.

The \textbf{breastw} dataset is also derived from the ``Breast Cancer Wisconsin (Original)`` dataset.
This is a 9-dimensional dataset containing 683 instances of which 239 represent malignant tumors and are treated as the outlier class.

The \textbf{cardio} dataset is derived from the ``Cardiotocography'' dataset.
The dataset is composed of measurements of fetal heart rate and uterine contraction features on cardiotocograms.
The are each labeled ``normal'', ``suspect'', and ``pathologic'' by expert obstetricians.
For anomaly detection, the ``normal'' class forms the inliers, the ``suspect'' class is discarded, and the ``pathologic'' class is downsampled to 176 instances forming the outliers.
This leaves us with 1831 instances with 21 attributes in the dataset.

The \textbf{cover} dataset is derived from the ``Covertype'' dataset.
The original dataset contains 581,012 instances with 54 attributes.
The dataset is used to predict the type of forest cover solely from cartographic variables.
The instances are labeled into seven different classes.
For outlier detection, we use only the 10 quantitative attributes as the features.
We treat class 2 (lodgepole pine) as the inliers, and class 4 (cottonwood/willow) as the outliers.
The remaining classes are discarded.
This leaves us with a 10-dimensional dataset with 286,048 instances of which 2,747 are outliers.

The \textbf{glass} dataset is derived from the ``Glass Identification'' dataset.
The study of classification of types of glass was motivated by criminological investigations where glass fragments left at crime scenes were used as evidence.
This dataset contains 214 instances with 9 attributes.
While there are several different types of glass in this dataset, class 6 is a clear minority with only 9 instances and, as such, points in class 6 are treated as the outliers while all other classes are treated as inliers.

The \textbf{http} dataset is derived from the original ``KDD Cup 1999'' dataset.
It contains 41 attributes (34 continuous and 7 categorical) which are reduced to 4 attributes (service, duration, src\_bytes, dst\_bytes).
Only the ``service'' attribute is categorical, dividing the data into \{http, smtp, ftp, ftp\_data, others\} subsets.
Here, only the ``http'' data is used.
The values of the continuous attributes are centered around 0, so they have been log-transformed far away from 0.
The original data contains 3,925,651 attacks in 4,898,431 records.
This smaller dataset is created with only 2,211 attacks in 567,479 records.

The \textbf{ionosphere} dataset is derived from the ``Ionosphere'' dataset.
It consists of 351 instances with 34 attributes.
One of the attributes is always 0 and, so, is discarded, leaving us with a 33-dimensional dataset.
The data come from radar measurements of the ionosphere from a system located in Goose Bay, Labrador.
The data are classified into ``good'' if the radar returns evidence of some type of structure in the ionosphere, and ``bad'' otherwise.
The ``good'' class serves as the inliers and the ``bad'' class serves as the outliers.

The \textbf{lympho} dataset is derived from the ``Lymphography'' dataset.
The data contain 148 instances with 18 attributes.
The instances are labeled ``normal find'', ``metastases'', ``malign lymph'', and ``fibrosis''.
The two minority classes only contain a total of six instances, and are combined to form the outliers.
The remaining 142 instances form the inliers.

The \textbf{mammography} dataset is derived from the original ``Mammography'' dataset provided by Aleksandar Lazarevic.
Its goal is to use x-ray images of human breasts to find calcified tissue as an early sign of breast cancer.
As such, the ``calcification'' class is considered as the outlier class while the ``non-calcification'' class is the inliers.
We have 11,183 instances with 6 attributes, of which 260 are ``calcifications.''

The \textbf{mnist} dataset is derived from the classic ``MNIST'' dataset of handwritten digits.
Digit-zero is considered the inlier class while 700 images of digit-six are the outliers.
Furthermore, 100 pixels are randomly selected as features from the original 784 pixels.

The \textbf{musk} dataset is derived from its namesake in the UCIMLR\@.
It is created from molecules that have been classified by experts as ``musk'' or ``non-musk''.
The data are downsampled to 3,062 instances with 166 attributes.
The ``musk'' class forms the outliers while the ``non-musk'' class forms the inliers.

The \textbf{optdigits} dataset is derived from the ``Optical Recognition of Handwritten Digits'' dataset.
Digits 1--9 form the inliers while 150 samples of digit-zero form the outliers.
This gives us a dataset of 5,216 instances with 64 attributes.

The \textbf{pendigits} dataset is derived from the ``Pen-Based Recognition of Handwritten Digits'' dataset from the UCI Machine Learning Repository.
The original collection of handwritten samples is reduced to 6,870 points, of which 156 are outliers.

The \textbf{pima} dataset is derived from the ``Pima Indians Diabetes'' dataset.
The original dataset presents a binary classification problem to detect diabetes.
This subset was restricted to female patients at least 21 years old of Pima Indian heritage.

The \textbf{satellite} dataset is derived from the ``Statlog (Landsat Satellite)'' dataset.
The smallest three classes (2, 4, and 5) are combined to form the outlier class while the other classes are combined to form the inlier class.
The train and test subsets are combined to produce a of 6,435 instances with 36 attributes.

The \textbf{satimage-2} dataset is also derived from the ``Satlog (Landsat Satellite)'' dataset.
Class 2 is downsampled to 71 instances that are treated as outliers, while all other classes are combined to form an inlier class.
This gives us 5,803 instances with 36 attributes.

The \textbf{shuttle} dataset is derived from the ``Statlog (Shuttle)'' dataset.
There are seven classes in the original dataset.
Here, class 4 is discarded, class 1 is treated as the inliers and the remaining classes, which are comparatively small, are combined into an outlier class.
This gives us 49,097 instances with 9 attributes, of which 3,511 are outliers.

The \textbf{smtp} is also derived from the ``KDD Cup 1999'' dataset.
It is pre-processed in the same way as the \textbf{http} dataset, except that the ``smtp'' service subset is used.
This version of the dataset only contains 95,156 instances with 3 attributes, of which 30 instances are outliers.

The \textbf{thyroid} dataset is also derived from the ``Thyroid Disease'' dataset.
The attribute selection is the same as for the \textbf{annthyroid} dataset but only the 3,772 training instances are used in this version.
The ``hyperfunction'' class, containing 93 instances, is treated as the outlier class, while the other two classes are combined to form an inlier class.

The \textbf{vertebral} dataset is derived from the ``Vertebral Column'' dataset.
6 attributes are derived to represent the shape and orientation of the pelvis and lumbar spine.
These attributes are: pelvic incidence, pelvic tilt, lumbar lordosis angle, sacral slope, pelvic radius and grade of spondylolisthesis.
Each instance comes from a different patient.
The ``Abnormal (AB)'' class of 210 instances are used as inliers while the ``Normal (NO)'' class is downsampled to 30 instances to be used as outliers.

The \textbf{vowels} dataset is derived from the ``Japanese Vowels'' dataset.
The UCIMLR presents this data as a multivariate time series of nine speakers uttering two Japanese vowels.
For outlier detection, each frame of each time-series is treated as a separate point.
There are 12 features associated with each time series, and these translate as the attributes for each point.
Data from speaker 1, downsampled to 50 points, form the outlier class.
Speakers 6, 7, and 8 form the inlier class.
The rest of the points are discarded.
This leaves is with 1,456 points in 12 dimensions, of which 50 are outliers.

The \textbf{wbc} dataset is derived from the ``Wisconsin-Breast Cancer (Diagnostics)'' dataset.
The dataset records measurements for breast cancer cases.
The benign class is treated as the inlier class, while the malignant class is downsampled to 21 points and serves as the outlier class.
This leaves us with 278 points in 30 dimensions.

The \textbf{wine} dataset is a collection of results of a chemical analysis of several wines from a region in Italy.
The data contain 129 samples having 13 attributes, and divided into 3 classes.
Classes 2 and 3 form the inliers while class 1, downsampled to 10 instances, is the outlier class.

\begin{table*}[!t]
\renewcommand{\arraystretch}{1.25}
\caption{Datasets used for Benchmarks}
\label{supplement:table:datasets}
\centering
\begin{tabular}{|c|c|c|c|c|}
\hline
\textbf{Dataset} & \textbf{Cardinality} & \textbf{\# Dim.} & \textbf{\# Outliers} & \textbf{\% Outliers} \\
\hline
annthyroid & 7,200 & 6 & 534 & 7.42 \\
\hline
arrhythmia & 452 & 274 & 66 & 15 \\
\hline
breastw & 683 & 9 & 239 & 35 \\
\hline
cardio & 1,831 & 21 & 176 & 9.6 \\
\hline
cover & 286,048 & 10 & 2,747 & 0.9 \\
\hline
glass & 214 & 9 & 9 & 4.2 \\
\hline
http & 567,479 & 4 & 2,211 & 0.4 \\
\hline
ionosphere & 351 & 33 & 126 & 36 \\
\hline
lympho & 148 & 18 & 6 & 4.1 \\
\hline
mammography & 11,183 & 6 & 260 & 2.32 \\
\hline
mnist & 7603 & 100 & 700 & 9.2 \\
\hline
musk & 3,062 & 166 & 97 & 3.2 \\
\hline
optdigits & 5,216 & 64 & 150 & 3 \\
\hline
pendigits & 6,870 & 16 & 156 & 2.27 \\
\hline
pima & 768 & 8 & 268 & 35 \\
\hline
satellite & 6,435 & 36 & 2036 & 32 \\
\hline
satimage-2 & 5,803 & 36 & 71 & 1.2 \\
\hline
shuttle & 59,097 & 9 & 3,511 & 7 \\
\hline
smtp & 95,156 & 3 & 30 & 0.03 \\
\hline
thyroid & 3,772 & 6 & 93 & 2.5 \\
\hline
vertebral & 240 & 6 & 30 & 12.5 \\
\hline
vowels & 1,456 & 12 & 50 & 3.4 \\
\hline
wbc & 278 & 30 & 21 & 5.6 \\
\hline
wine & 129 & 13 & 10 & 7.7 \\
\hline
\end{tabular}
\end{table*}

\section{Complexity of CHAODA}
\label{supplement:sec:complexity-chaoda}

Here we provide short proofs for the time complexity and space complexity of the CHAODA algorithms.
For each algorithm, we have
a dataset $X$ with $n = |X|$ points and
a graph $G(V, E)$ of clusters/vertices $V$ and edges $E$ between overlapping clusters.

\subsection{CLAM Clustering}

We use CLAM to build the cluster-tree and the induced graphs.
The time complexity of clustering is the same as for clustering in CHESS~\cite{ishaq2019clustered}; i.e., expected $\mathcal{O}(nlogn)$ and worst-case $\mathcal{O}(n^2)$ where $n$ is the size of the dataset.

The cost for inducing graphs depends on whether it is a layer-graph or an optimal graph.
For both types of graphs, we first have to select the right clusters, and then find neighbors based on cluster overlap.

We implemented CLAM in Python and the language does not have tail-call optimization for recursive functions.
Therefore we implement partition to, instead of recursing until reaching leaves, iteratively increase the depth of the tree.
During the course of this partition, we store a map from tree-depth to a set of clusters at that depth.
Therefore, selecting all cluster at a given depth costs $\mathcal{O}(1)$ time and $\mathcal{O}(|V|)$ space where $V$ is the set of selected clusters.
Selecting clusters for optimal-graphs is more expensive.
First, we use a trained meta-ml model to predict the AUC contribution from each cluster in a tree; this costs $\mathcal{O}(n)$ time and $\mathcal{O}(n)$ space.
Next, we sort the clusters by this predicted value; this costs $\mathcal{O}(nlogn)$ time and $\mathcal{O}(n)$ space.
Finally, we perform a linear pass over the clusters to select the best for the graph, while discarding the ancestors and descendants of any cluster that has already been selected; this costs $\mathcal{O}(n)$ time and $\mathcal{O}(|V|)$ space.
Therefore, the total cost of selecting clusters for optimal graphs is $\mathcal{O}(nlogn)$ time and $\mathcal{O}(n)$ space.

Once the clusters have been selected for a graph, we have to find every pair of clusters with overlapping volumes.
Na\"ively, this can be done with an all-pairs distance computation for a cost of $\mathcal{O}(|V|^2)$ for time and space.
However, our implementation is superior to the na\"ive method although the proof is beyond the scope of this supplement.

\subsection{Relative Cluster Cardinality}

This algorithm performs a single linear pass over the vertices in the graph.
The cardinalities of clusters are cached during the tree-building phase of clam.
Each lookup from this cache costs $\mathcal{O}(1)$.
For a graph $G(V, E)$ the time-complexity is trivially $\mathcal{O}(|V|)$.
Since each cluster stores its cardinality, the space complexity is also $\mathcal{O}(|V|)$.

\subsection{Relative Component Cardinality}

This method first finds the components of the graph.
This costs $\mathcal{O}(|E|)$ time because we have to check each edge once.
The cardinality of each component is cached when traversing the clusters to find components, thus the space complexity is $\mathcal{O}(|C|)$ where $C$ is the set of distinct connected components in the graph.
With this done, the algorithm performs a single linear pass over each component.
This brings the total worst-case cost to $\mathcal{O}(|E| + |V|)$.

\subsection{Graph Neighborhood}

This algorithm performs a linear pass over the clusters in the graph and first computes the eccentricity of each cluster.
Finding the eccentricity of a vertex in a graph is worst-case $\mathcal{O}(|E|)$ time when the graph consists of a single component.
This brings the total cost up to $\mathcal{O}(|E| \cdot |V|)$, with space complexity $\mathcal{O}(|V|+|E|)$.
Next, the algorithm performs a traversal from each cluster.
This adds a constant factor of $\mathcal{O}(|E|)$ to the time complexity and $\mathcal{O}(|V|)$ to the space complexity, which can be ignored.
The total time-complexity of this algorithm is thus $\mathcal{O}(|E| \cdot |V|)$ and its space complexity is $\mathcal{O}(|V|)$ (because only the size of each graph-neighborhood needs to be stored).

\subsection{Child-Parent Cardinality Ratio}

While building the tree with CLAM, we cache the child-parent cardinality ratios of every cluster, because it proved useful for purposes other than anomaly detection.
This method performs a single linear pass over the clusters in the graph and looks-up the cached child-parent ratios as needed.
The time-complexity is thus $\mathcal{O}(|V|)$.
Since the ratios are cached with their respective clusters, the space complexity is $\mathcal{O}(|V|)$.

\subsection{Stationary Probabilities}

This method starts by computing a transition matrix for each component in the graph.
We set the transition probability from a cluster to a neighbor to be inversely proportional to the distance between their centers, normalized by all possible neighbors of the cluster.
We successively square this matrix until it converges.
The transition matrices from our graphs obey the criteria required for convergence as proven in~\cite{levin2017markov}.
Matrix multiplication for square matrices costs $\mathcal{O}(|V|^{2.373})$ with the Coppersmith-Winograd algorithm~\cite{coppersmith1987matrix}.
Thus the worst-case time complexity is the same as that for the matrix-multiplication algorithm employed.
For space, we need only store a single $|V| \times |V|$ matrix, giving us a space complexity of $\mathcal{O}(|V|^2)$.

In practice, $|V| \ll n$ and graphs only rarely consist of only one component.
Thus, the average run-time performance is much better than that suggested by the quadratic space time-complexity.

\subsection{Vertex Degree}

Since we already have a graph with vertices and edges, calculating the degree of each vertex only costs $\mathcal{O}(1)$ time.
Thus, the complexity of this algorithm is $\mathcal{O}(|V|)$.

\subsection{Normalization}

Normalizing the outlier scores requires finding the mean and standard deviation of the raw scores, followed by a linear pass over the set of scores.
Thus the time-complexity of this step is $\mathcal{O}(n)$.
Since we need to store a score for each point, the space complexity is $\mathcal{O}(n)$.

The algorithm is presented in Algorithm~\ref{alg:normalization}.

\begin{algorithm}[h]
    \caption{Gaussian Normalization}
    \label{alg:normalization}
\begin{algorithmic}[1]
    \REQUIRE $X$, a dataset
    \REQUIRE $S$, a set of outlier scores for each point in $X$
    \STATE $erf: x \mapsto \frac{2}{\sqrt{\pi}} \int_{0}^{x} e^{-u^2} \,du $
    \STATE $\mu \gets mean(S)$
    \STATE $\sigma \gets std(S)$
    \FOR {point $p \in X$}
        \STATE $S[p] \gets \frac{1}{2} \Big( 1 + erf \big(\frac{S[p] - \mu}{\sigma \cdot \sqrt{2}}\big) \Big) $
    \ENDFOR
\end{algorithmic}
\end{algorithm}

\subsection{Ensemble}

Given the normalized scores from the individual methods, we combine the scores by voting among them in an ensemble.
There is a small, constant, number of scores for each point; each score is from a different graph built using the meta-ml models.
We simply take the mean of all scores for each point.
Thus the time-complexity of voting among these scores is $\mathcal{O}(n)$ for the entire dataset.
Since we need to store a score for each point, the space complexity is $\mathcal{O}(n)$.

\section{UMAP Visualization}
\label{supplement:sec:umap-visualization}

\begin{figure*}
   \centering
   \includegraphics[width=2.5in]{images/umaps/cardio-euclidean-umap2d.png}
   \includegraphics[width=2.5in]{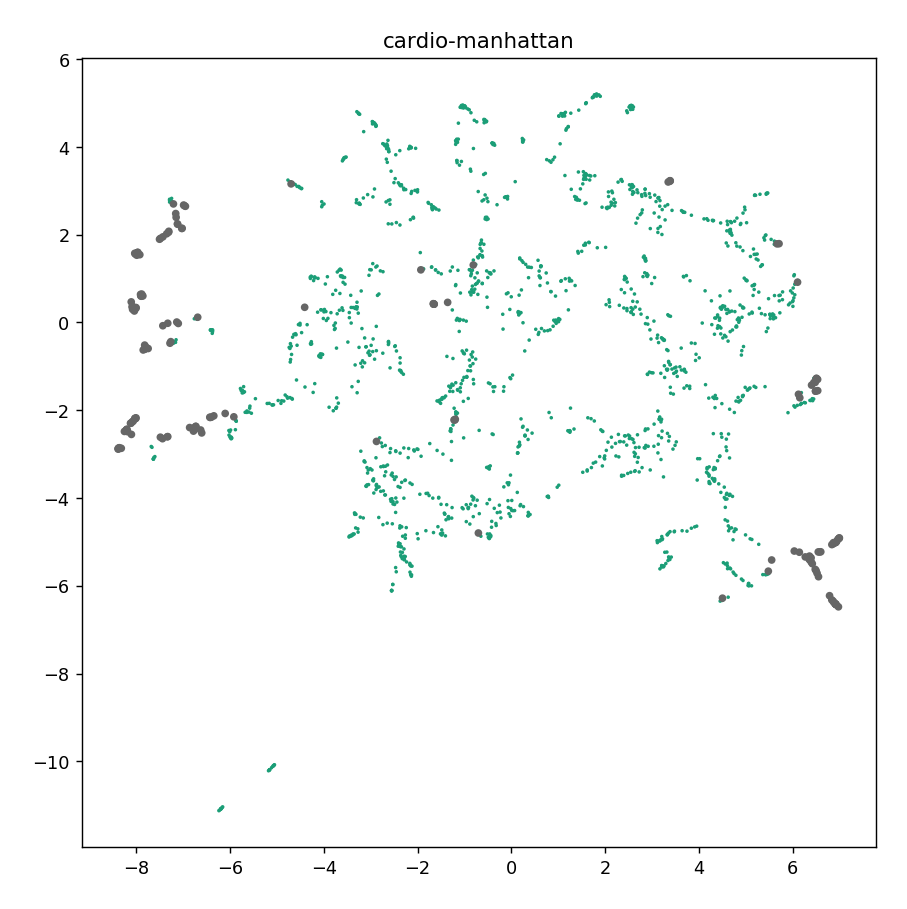}
   \includegraphics[width=2.5in]{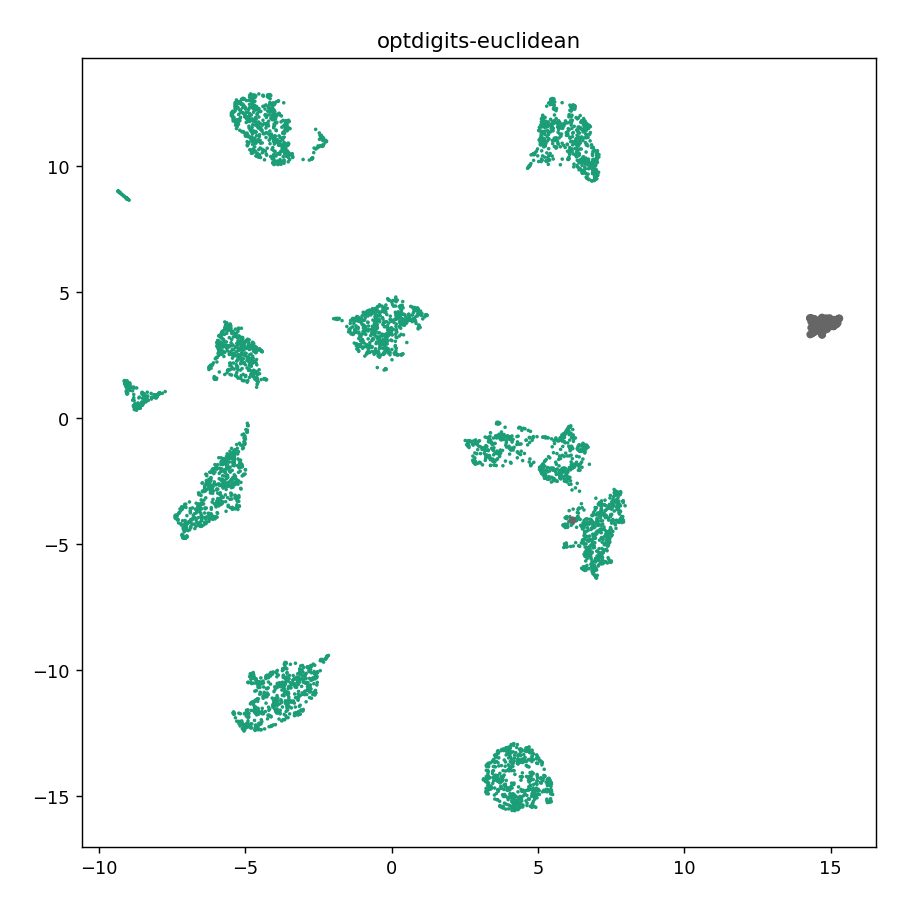}
   \includegraphics[width=2.5in]{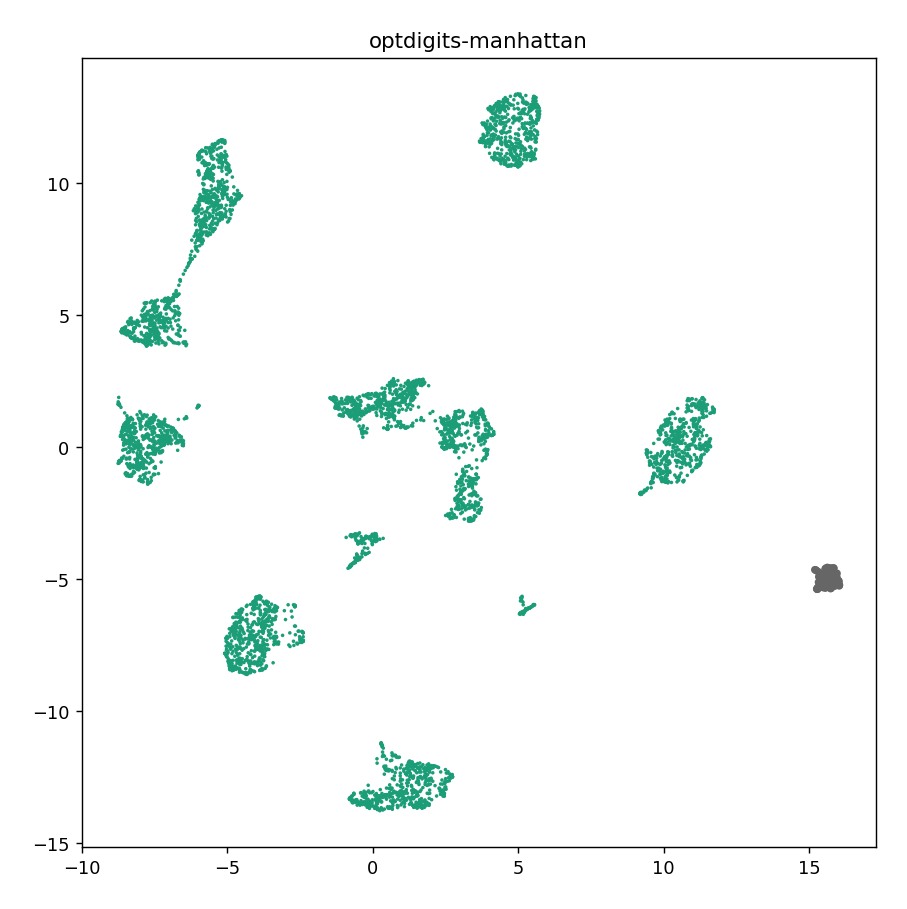}
   \includegraphics[width=2.5in]{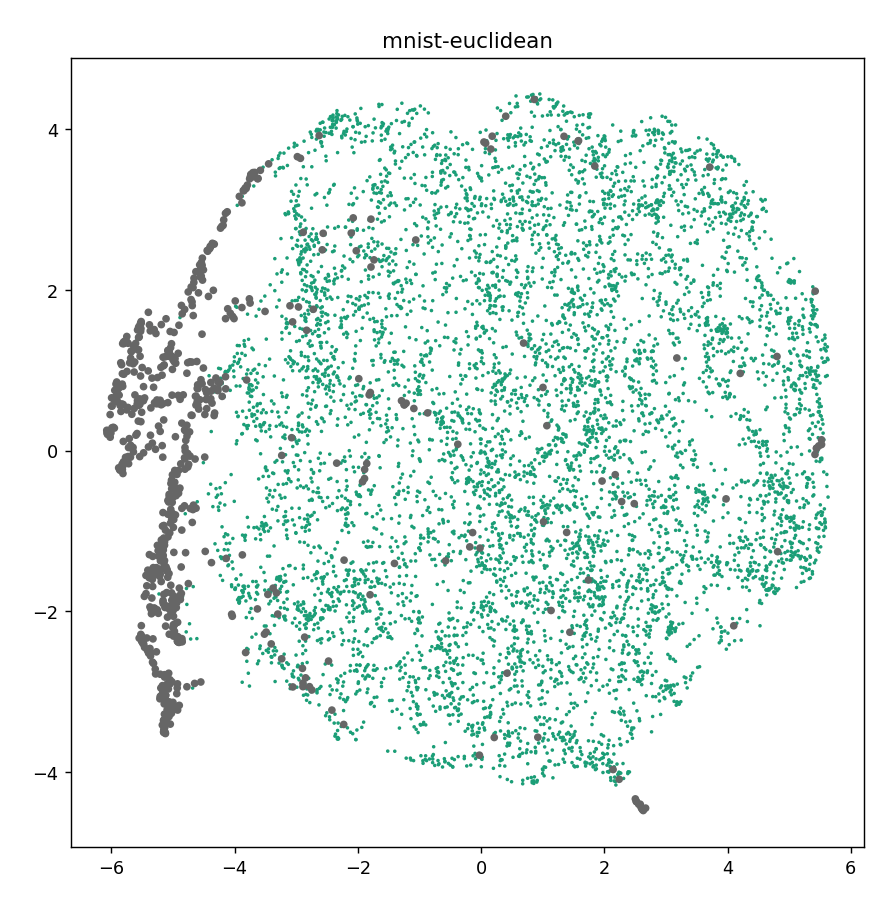}
   \includegraphics[width=2.5in]{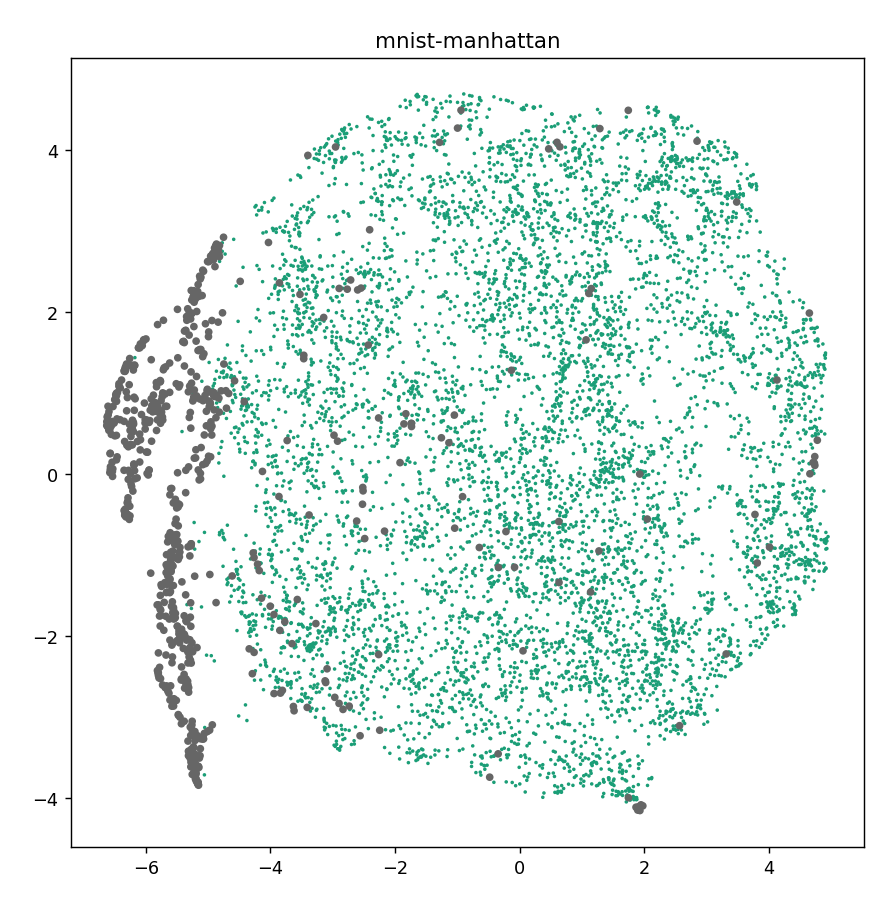}
   \caption{UMAP projections of Cardio (first row), Optdigits (second row) and Mnist (third row).
   The distance metrics used are Euclidean (left column) and Manhattan (right column).
   Anomalies are in gray.
   Note that for MNIST, the UMAP projection does not find much structure, though most of the anomalies congregate to one side.
   For Cardio, there is a single main component to the manifold, and anomalies tend to be at the edges of that manifold.
   For OptDigits, there are several distinct pieces to the manifold, perhaps corresponding to different digits.
   Most algorithms perform comparably on MNIST, while CHAODA outperforms others on Cardio and OptDigits.}
   \label{supplement:fig:conclusions:umap-embeddings-1}
\end{figure*}

A visualization in Figure~\ref{supplement:fig:conclusions:umap-embeddings-1} using UMAP illustrates a handful of different examples;
the anomalies in the Cardio and OptDigits datasets, where CHAODA outperforms other methods, appear to be at the edges of a complex manifold (though, clearly, the UMAP projection has distorted the manifold).
In the Mnist dataset, where several methods perform fairly well, the distribution is less interesting.
Most anomalies are off to one side but there are several interspersed among the inliers.

\begin{figure*}
   \centering
   \includegraphics[width=2.5in]{images/umaps/pima-euclidean-umap2d.png}
   \includegraphics[width=2.5in]{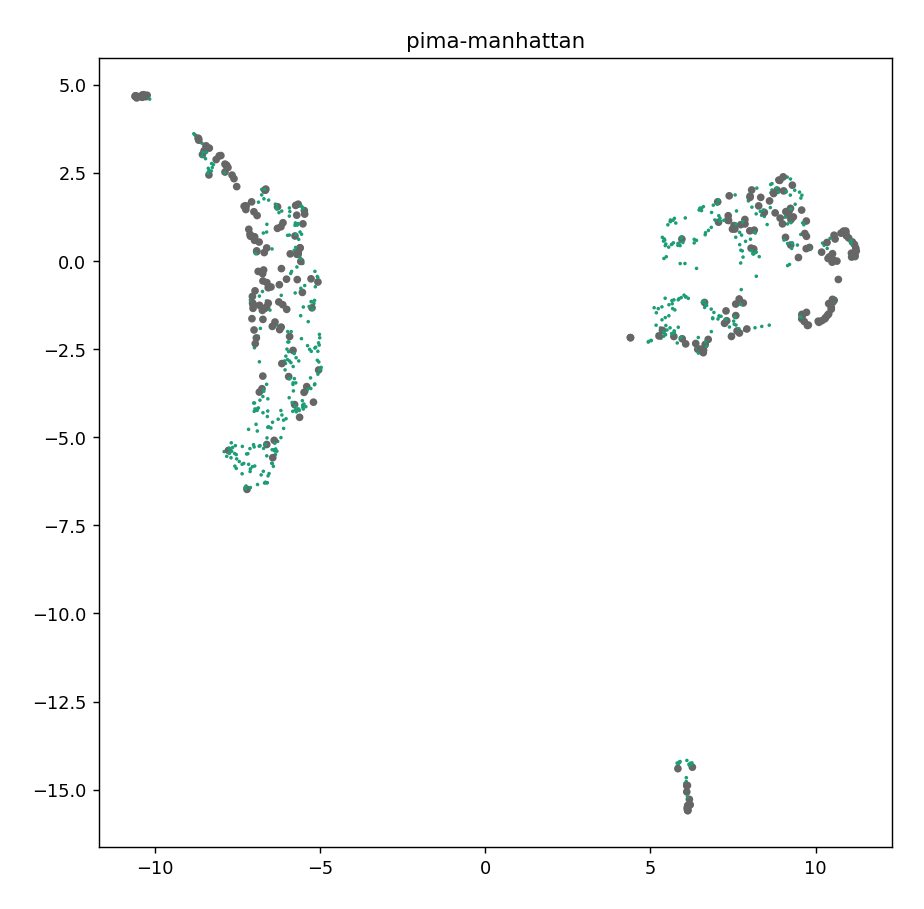}
   \caption{UMAP projections of the Pima dataset.
   All algorithms performed poorly on Pima.
   This may be because the anomalies and the outliers seem inseparable in the projection.}
   \label{supplement:fig:conclusions:umap-embeddings-2}
\end{figure*}

In Figure~\ref{supplement:fig:conclusions:umap-embeddings-2}, we show UMAP visualizations of the Pima dataset.
The inliers and outliers seem inseparable, and so all the methods perform poorly.

\section{Performance}
\label{supplement:sec:performance}

\subsection{Run-time performance on Test set of Datasets}

Tables~\ref{supplement:table:results:test-time-1}~and~\ref{supplement:table:results:test-time-2} report the running time, in seconds, of CHAODA and each competitor.
The fastest methods on each dataset are presented in bold face.

\subsection{AUC and Runtime performance on Train set of Datasets}

Tables~\ref{supplement:table:results:train-performance}~and~\ref{supplement:table:results:train-time} report the  the AUC performance and running time, respectively, of CHAODA and each competitor on the train set of datasets.

\subsection{Performance of Individual CHAODA Algorithms}

The ensemble of CHAODA algorithms is discussed extensively in the main paper, but we did not have room to discuss or present the performance of the individual algorithms.
Due to the large numbers of graphs generated for the ensemble and with each method being applied to each graph, we cannot provide these intermediate results as a table in this document.
We instead provide a .csv file which will be available for download.

\subsection{Performance on APOGEE-2}

For a read ``Big-Data'' challenge, we applied CHAODA to the APOGEE-2 dataset from the SDSS.
This dataset contains spectra of stars, i.e.\ intensity/flux measured at several wavelengths, in $8,757$ dimensions.
We extracted $528,323$ spectra from data-release 15.
CHAODA was able to produce anomaly scores for these data in approximately 2 hours and 36 minutes.
These scores, and the associated metadata, are provided in our github repository in the `sdss' directory.

\begin{table*}[!t]
\renewcommand{\arraystretch}{1.15}
\caption{Time taken, in seconds, on the first half of the Test Datasets}
\label{supplement:table:results:test-time-1}
\vskip 0.15in
\begin{center}
\begin{small}
\begin{tabular}{|c|c|c|c|c|c|c|c|c|c|}
\hline
\textbf{Model} & \textbf{Arrhy} & \textbf{BreastW} & \textbf{Cardio} & \textbf{Cover} & \textbf{Glass} & \textbf{Http} & \textbf{Iono} & \textbf{Lympho} & \textbf{Mammo} \\
\hline
CHAODA-fast    & 8.06           & 5.01             & 42.37           & 829.62         & 1.73           & 5e3           & 3.71          & 1.78            & 29.92          \\
\hline
CHAODA         & 27.43          & 14.00            & 344.27          & 6e3            & 7.44           & 2e4           & 22.13         & 1.74            & 244.81         \\
\hline
ABOD &                0.34 &             0.20 &            0.72 &          24.02 &  \textbf{0.07} &         19.08 &                0.13 &   \textbf{0.05} &           3.82 \\
\hline
AutoEncoder &                8.00 &             6.18 &            9.05 &         183.70 &           3.99 &        154.20 &                4.99 &            3.79 &          35.26 \\
\hline
CBLOF &                0.16 &             0.13 &            0.17 &    \textit{EX} &  \textbf{0.06} &   \textit{EX} &       \textbf{0.07} &   \textbf{0.05} &           0.20 \\
\hline
COF &                1.24 &             1.84 &           12.63 &       2e4 &           0.24 &      2e5 &                0.60 &            0.14 &         513.46 \\
\hline
HBOS &       \textbf{0.08} &    \textbf{0.00} &   \textbf{0.01} &           1.13 &  \textbf{0.00} & \textbf{0.01} &       \textbf{0.01} &   \textbf{0.01} &  \textbf{0.01} \\
\hline
IFOREST &                0.43 &             0.34 &            0.43 &           4.61 &           0.30 &          3.95 &                0.33 &            0.30 &           0.95 \\
\hline
KNN &                0.19 &    \textbf{0.07} &            0.30 &          11.46 &  \textbf{0.02} &          7.47 &       \textbf{0.05} &   \textbf{0.02} &           1.58 \\
\hline
LMDD &               14.85 &             2.62 &           22.12 &       1e4 &           0.57 &       4e3 &                1.73 &            0.42 &         243.01 \\
\hline
LOCI &              307.04 &      \textit{TO} &     \textit{TO} &    \textit{TO} &          25.68 &   \textit{TO} &              120.92 &            9.78 &    \textit{TO} \\
\hline
LODA &       \textbf{0.04} &    \textbf{0.03} &   \textbf{0.05} &           0.81 &  \textbf{0.03} &          0.66 &       \textbf{0.03} &   \textbf{0.03} &           0.17 \\
\hline
LOF &                0.16 &    \textbf{0.01} &            0.18 &          10.24 &  \textbf{0.00} &          1.93 &       \textbf{0.01} &   \textbf{0.00} &           0.59 \\
\hline
MCD &                5.08 &             0.64 &            0.89 &          28.65 &  \textbf{0.05} &          8.21 &                0.15 &   \textbf{0.05} &           2.11 \\
\hline
MOGAAL &               46.46 &            42.09 &          116.73 &    \textit{TO} &          40.67 &   \textit{TO} &               40.86 &           37.84 &    \textit{TO} \\
\hline
OCSVM &       \textbf{0.10} &    \textbf{0.02} &            0.23 &         257.95 &  \textbf{0.00} &        257.28 &       \textbf{0.01} &   \textbf{0.00} &           6.37 \\
\hline
SOD &                1.19 &             1.89 &           13.39 &    \textit{TO} &           0.31 &   \textit{TO} &                0.76 &            0.20 &         521.25 \\
\hline
SOGAAL &                6.70 &             5.03 &           14.43 &         591.14 &           3.95 &        597.17 &                4.71 &            3.96 &          92.27 \\
\hline
SOS &                0.69 &            47.40 &            4.11 &    \textit{TO} &           0.18 &   \textit{TO} &                0.33 &            0.11 &    \textit{TO} \\
\hline
VAE &               10.00 &             7.75 &           10.89 &         223.09 &           5.27 &        175.59 &                6.66 &            5.21 &          40.87 \\
\hline
\end{tabular}
\end{small}
\end{center}
\vskip -0.1in
\end{table*}

\begin{table*}[!t]
\renewcommand{\arraystretch}{1.15}
\caption{Time taken, in seconds, on the second half of the Test Datasets}
\label{supplement:table:results:test-time-2}
\vskip 0.15in
\begin{center}
\begin{small}
\begin{tabular}{|c|c|c|c|c|c|c|c|c|c|}
\hline
\textbf{Model} & \textbf{Musk} & \textbf{OptDigits} & \textbf{Pima} & \textbf{SatImg-2} & \textbf{Smtp} & \textbf{Vert} & \textbf{Vowels} & \textbf{WBC} & \textbf{Wine} \\
\hline
CHAODA-fast    & 40.22         & 75.81              & 8.16          & 58.63             & 156.97        & 2.59          & 29.96           & 4.71          & 0.89          \\
\hline
CHAODA         & 601.12        & 8e3                & 24.50         & 4e3               & 413.86        & 2.32          & 214.64          & 7.21          & 1.08          \\
\hline
ABOD &          4.39 &               6.81 &          0.26 &                3.20 &         18.00 &      \textbf{0.08} &            0.51 &          0.14 & \textbf{0.04} \\
\hline
AutoEncoder &         23.40 &              24.29 &          4.65 &               23.73 &        195.61 &               2.85 &            6.98 &          4.70 &          3.31 \\
\hline
CBLOF &          0.27 &               0.44 & \textbf{0.08} &                0.31 &          0.65 &      \textbf{0.06} &   \textbf{0.10} & \textbf{0.08} & \textbf{0.05} \\
\hline
COF &         45.60 &             119.57 &          2.40 &              141.84 &      2e4 &               0.28 &            8.06 &          0.68 &          0.11 \\
\hline
HBOS & \textbf{0.07} &      \textbf{0.04} & \textbf{0.00} &       \textbf{0.02} & \textbf{0.01} &      \textbf{0.00} &   \textbf{0.01} & \textbf{0.01} & \textbf{0.00} \\
\hline
IFOREST &          0.93 &               0.92 &          0.35 &                0.79 &          3.90 &               0.30 &            0.40 &          0.33 &          0.29 \\
\hline
KNN &          3.50 &               5.53 & \textbf{0.09} &                1.84 &          6.83 &      \textbf{0.03} &            0.18 & \textbf{0.05} & \textbf{0.01} \\
\hline
LMDD &        375.11 &             421.22 &          2.86 &              302.94 &       4e3 &               0.59 &            9.47 &          1.91 &          0.33 \\
\hline
LOCI &   \textit{TO} &        \textit{TO} &   \textit{TO} &         \textit{TO} &   \textit{TO} &              35.97 &     \textit{TO} &        154.49 &          6.84 \\
\hline
LODA & \textbf{0.07} &      \textbf{0.10} & \textbf{0.04} &       \textbf{0.10} &          0.73 &      \textbf{0.03} &   \textbf{0.05} & \textbf{0.04} & \textbf{0.03} \\
\hline
LOF &          3.59 &               5.49 & \textbf{0.02} &                1.50 &          1.36 &      \textbf{0.00} &   \textbf{0.06} & \textbf{0.01} & \textbf{0.00} \\
\hline
MCD &         84.74 &               7.24 &          0.69 &                6.83 &         13.78 &      \textbf{0.05} &            0.81 &          0.11 & \textbf{0.05} \\
\hline
MOGAAL &        267.93 &             415.02 &         40.89 &              463.18 &   \textit{TO} &              39.86 &           80.35 &         40.65 &         38.10 \\
\hline
OCSVM &          2.51 &               3.71 & \textbf{0.03} &                3.14 &        252.52 &      \textbf{0.00} &            0.11 & \textbf{0.01} & \textbf{0.00} \\
\hline
SOD &         56.94 &             131.49 &          2.48 &              210.39 &   \textit{TO} &               0.54 &           13.92 &          1.03 &          0.19 \\
\hline
SOGAAL &         29.55 &              44.72 &          5.20 &               48.79 &        592.27 &               4.69 &           10.38 &          4.76 &          4.01 \\
\hline
SOS &          9.37 &              26.09 &          1.01 &               34.96 &   \textit{TO} &               0.21 &            2.81 &          0.36 & \textbf{0.10} \\
\hline
VAE &         30.38 &              30.58 &          5.67 &               30.41 &        176.84 &               4.03 &           10.46 &          5.33 &          4.62 \\
\hline
\end{tabular}
\end{small}
\end{center}
\vskip -0.1in
\end{table*}

\begin{table*}[!t]
\renewcommand{\arraystretch}{1.25}
\caption{Performance on Train Datasets}
\label{supplement:table:results:train-performance}
\vskip 0.15in
\begin{center}
\begin{small}
\begin{tabular}{|c|c|c|c|c|c|c|}
\hline
\textbf{Model} & \textbf{Annthy} & \textbf{Mnist} & \textbf{PenDigits} & \textbf{Satellite} & \textbf{Shuttle} & \textbf{Thyroid} \\
\hline
CHAODA-fast    & 0.64            & 0.67           & 0.83               & 0.63               & \textbf{0.97}    & \textbf{0.89}    \\
\hline
CHAODA         & \textbf{0.85}   & \textbf{0.71}  & \textbf{0.87}      & \textbf{0.77}      & 0.86             & \textbf{0.91}     \\
\hline
ABOD &                0.50 &           0.60 &               0.53 &               0.51 &             0.54 &             0.50 \\
\hline
AutoEncoder &                0.69 &           0.67 &               0.58 &               0.63 &             0.94 &             0.88 \\
\hline
CBLOF &                0.59 &           0.62 &               0.59 &               0.68 &    \textbf{0.99} &             0.87 \\
\hline
COF &                0.59 &           0.56 &               0.53 &               0.56 &             0.52 &             0.49 \\
\hline
HBOS &       \textbf{0.84} &           0.53 &               0.52 &               0.62 &             0.74 &             0.86 \\
\hline
IFOREST &                0.70 &           0.61 &               0.63 &               0.70 &             0.91 &    \textbf{0.91} \\
\hline
KNN &                0.65 &           0.65 &               0.51 &               0.56 &             0.53 &             0.56 \\
\hline
LMDD &                0.52 &           0.59 &               0.56 &               0.42 &             0.92 &             0.70 \\
\hline
LOCI &         \textit{TO} &    \textit{TO} &        \textit{TO} &        \textit{TO} &      \textit{TO} &      \textit{TO} \\
\hline
LODA &                0.63 &           0.66 &               0.57 &               0.65 &             0.96 &    \textbf{0.90} \\
\hline
LOF &                0.60 &           0.57 &               0.52 &               0.57 &             0.53 &             0.49 \\
\hline
MCD &                0.72 &           0.57 &               0.53 &               0.58 &             0.96 &             0.85 \\
\hline
MOGAAL &                0.46 &    \textit{TO} &               0.67 &               0.59 &      \textit{TO} &             0.49 \\
\hline
OCSVM &                0.62 &           0.63 &               0.59 &               0.62 &    \textbf{0.97} &             0.78 \\
\hline
SOD &                0.64 &           0.55 &               0.52 &               0.52 &      \textit{TO} &             0.53 \\
\hline
SOGAAL &                0.46 &           0.57 &               0.57 &               0.60 &             0.96 &             0.49 \\
\hline
SOS &                0.50 &           0.52 &               0.52 &               0.47 &      \textit{TO} &             0.50 \\
\hline
VAE &                0.69 &           0.67 &               0.58 &               0.63 &             0.94 &             0.88 \\
\hline
\end{tabular}
\end{small}
\end{center}
\vskip -0.1in
\end{table*}

\begin{table*}[!t]
\renewcommand{\arraystretch}{1.25}
\caption{Time taken, in seconds, on Train Datasets}
\label{supplement:table:results:train-time}
\centering
\begin{tabular}{|c|c|c|c|c|c|c|}
\hline
\textbf{Model} & \textbf{Annthy} & \textbf{Mnist} & \textbf{PenDigits} & \textbf{Satellite} & \textbf{Shuttle} & \textbf{Thyroid} \\
\hline
CHAODA-fast    & 35.92           & 93.81          & 58.03              & 54.60              & 17.07            & 35.51           \\
\hline
CHAODA         & 112.86          & 732.15         & 481.12             & 314.60             & 1e4              & 108.99           \\
\hline
ABOD &                3.42 &          16.39 &               2.83 &               3.53 &            21.58 &             1.25 \\
\hline
AutoEncoder &               22.80 &          40.54 &              23.87 &              26.74 &           158.72 &            12.84 \\
\hline
CBLOF &                1.01 &           1.14 &               0.23 &               0.30 &             0.71 &             0.21 \\
\hline
COF &              215.63 &         277.91 &             199.03 &             176.35 &         1e4 &            54.82 \\
\hline
HBOS &                1.58 &  \textbf{0.06} &      \textbf{0.01} &      \textbf{0.02} &    \textbf{0.03} &    \textbf{0.00} \\
\hline
IFOREST &                0.73 &           1.41 &               0.80 &               0.84 &             3.24 &             0.54 \\
\hline
KNN &                0.84 &          14.30 &               1.25 &               2.05 &             9.13 &             0.42 \\
\hline
LMDD &              107.05 &        1e3 &             197.48 &             371.63 &          6e3 &            32.69 \\
\hline
LOCI &         \textit{TO} &    \textit{TO} &        \textit{TO} &        \textit{TO} &      \textit{TO} &      \textit{TO} \\
\hline
LODA &                0.13 &           0.13 &               0.12 &               0.11 &             0.59 &    \textbf{0.08} \\
\hline
LOF &                0.26 &          14.79 &               0.90 &               1.71 &             6.30 &    \textbf{0.09} \\
\hline
MCD &                1.50 &          20.56 &               3.01 &              10.39 &            10.78 &             1.03 \\
\hline
MOGAAL &              594.69 &    \textit{TO} &             561.40 &             513.84 &      \textit{TO} &           280.80 \\
\hline
OCSVM &                2.70 &          11.30 &               3.04 &               3.83 &           162.18 &             0.68 \\
\hline
SOD &              262.80 &         281.86 &             195.81 &             253.71 &      \textit{TO} &            91.48 \\
\hline
SOGAAL &               61.14 &          68.03 &              55.40 &              50.07 &           460.66 &            31.43 \\
\hline
SOS &              101.82 &          54.22 &              46.02 &              42.62 &      \textit{TO} &            42.10 \\
\hline
VAE &               27.95 &          50.16 &              29.63 &              33.51 &           182.23 &            16.18 \\
\hline
\end{tabular}
\end{table*}


\end{document}